\documentclass[sigconf]{acmart}

\usepackage{threeparttable} 

\AtBeginDocument{%
  \providecommand\BibTeX{{%
    \normalfont B\kern-0.5em{\scshape i\kern-0.25em b}\kern-0.8em\TeX}}}

\setcopyright{none}
\acmYear{2026}
\copyrightyear{2026}

\acmConference[arXiv]{arXiv Preprint}{2026}{}
\renewcommand\footnotetextcopyrightpermission[1]{}
\graphicspath{{./images/}} 

\usepackage{tcolorbox}
\usepackage{enumitem}
\usepackage{tabularx}
\usepackage{float}

\begin{document}

\title{LAMUS: A Large-Scale Corpus for Legal Argument Mining from U.S. Caselaw using LLMs}

\author{Serene Wang}
\affiliation{%
  \institution{University of North Texas}
  \city{Denton}
  \state{Texas}
  \country{USA}
  \postcode{76201}
}
\email{SereneWang@my.unt.edu}
\orcid{1234-5678-9012}

\author{Lavanya Pobbathi}
\affiliation{%
  \institution{University of North Texas}
  \city{Denton}
  \state{Texas}
  \country{USA}
  \postcode{76201}
}
\email{LavanyaPobbathi@my.unt.edu}
\orcid{0009-0009-1731-0543}

\author{Haihua Chen}
\authornote{Corresponding author.}
\affiliation{%
  \institution{University of North Texas}
  \city{Denton}
  \state{Texas}
  \country{USA}
  \postcode{76201}
}
\email{haihua.chen@unt.edu}
\orcid{0000-0002-7088-9752}

\renewcommand{\shortauthors}{Wang, Pobbathi and Chen}

\begin{abstract}
Legal argument mining aims to identify and classify the functional components of judicial reasoning, such as facts, issues, rules, analysis, and conclusions. Progress in this area is limited by the lack of large-scale, high-quality annotated datasets for U.S. caselaw, particularly at the state level. This paper introduces LAMUS, a sentence-level legal argument mining corpus constructed from U.S. Supreme Court decisions and Texas criminal appellate opinions. The dataset is created using a data-centric pipeline that combines large-scale case collection, LLM-based automatic annotation, and targeted human-in-the-loop quality refinement.
We formulate legal argument mining as a six-class sentence classification task and evaluate multiple general-purpose and legal-domain language models under zero-shot, few-shot, and chain-of-thought prompting strategies, with LegalBERT as a supervised baseline. Results show that chain-of-thought prompting substantially improves LLM performance, while domain-specific models exhibit more stable zero-shot behavior. LLM-assisted verification corrects nearly 20\% of annotation errors, improving label consistency. Human verification achieves Cohen's Kappa $\kappa = 0.85$, confirming annotation quality. LAMUS provides a scalable resource and empirical insights for future legal NLP research. All code and datasets can be accessed for reproducibility in GitHub at: \url{https://github.com/LavanyaPobbathi/LAMUS/tree/main}.
\end{abstract}




\maketitle

\section{Introduction}


Argument mining aims to automatically identify and structure argumentative components in text, enabling downstream reasoning and decision-support tasks \cite{chen2022construction}. In the legal domain, argument mining encompasses the extraction and classification of facts, issues, rules, legal analysis, and conclusions, which together form the basis of judicial reasoning. Automating these processes has the potential to significantly improve legal research efficiency, support judicial decision-making, and advance AI-driven legal intelligence \cite{mochales2011argumentation, zhang2022enhancing}. However, progress in legal argument mining for U.S. caselaw remains constrained by the scarcity of large-scale, high-quality annotated corpora, particularly at the state level. Existing datasets either focus on non-U.S. jurisdictions or are limited in size and scope, restricting systematic evaluation of large language models (LLMs) for modeling U.S. judicial reasoning.

Legal argument mining presents unique challenges due to the complexity of legal language, hierarchical reasoning structures, and extensive references to precedent and statutory interpretation \cite{ariai2025natural, siino2025exploring}. Recent work also highlights the importance of improving the logical reasoning capabilities of large language models when applied to legal analysis and decision-making tasks \cite{kant2025towards}. Judicial opinions rely on specialized language, hierarchical reasoning, and extensive references to precedent and statute, making both annotation and modeling substantially more complex than in general-domain argument mining. Typical workflows involve argument component detection, argument classification, and, in more advanced settings, reconstruction of argumentative structures and reasoning chains. The rapid development of large language models has led to increasing interest in evaluating their performance in specialized legal tasks and workflows \cite{guha2023legalbench, li2025legalagentbench}.


Recent research has advanced fine-grained extraction tasks, particularly sentence- or span-level classification of argumentative roles such as claims, premises, and counterarguments. Natural language processing techniques are increasingly applied to legal documents for tasks such as judgment prediction, argument mining, and legal question answering \cite{siino2025exploring}. Much of this work has focused on European Court of Human Rights (ECHR) cases \cite{poudyal2020echr} or Chinese legal corpora such as CAIL \cite{xiao2018cail2018}. Benchmarks such as LegalBench further explore higher-level reasoning tasks and evaluate the performance of LLMs in legal settings \cite{guha2023legalbench, li2025legalagentbench}, while recent work has also explored the development of domain-specific legal language models tailored to the U.S. legal system \cite{shu2024lawllm}. Although LLMs demonstrate strong generalization and flexibility, prior studies suggest that fine-tuned transformer models pretrained on legal corpora often remain more reliable and interpretable for classification tasks, especially under class imbalance and strict faithfulness requirements \cite{chalkidis-etal-2020-legal, ji2023survey}. Importantly, existing richly annotated resources such as ECHR-AM and LAM:ECHR are limited to non-U.S. jurisdictions. Despite the centrality of U.S. caselaw in legal research and practice, large-scale, systematically annotated U.S. argument mining corpora remain scarce, particularly at the state level.


To address these gaps, we introduce LAMUS, a large-scale legal argument mining corpus constructed from U.S. Supreme Court decisions and Texas criminal appellate opinions. The dataset contains judicial opinions from 1921-2025 and approximately 2,900,083  sentences, each annotated with one of six argumentative roles: Fact, Issue, Rule/Law/Holding, Analysis, Conclusion, or Other. Building on prior work in semi-automated corpus construction \cite{chen2022construction}, we adopt a data-centric approach that integrates corpus acquisition, automatic annotation using LLMs, targeted quality control, and systematic model evaluation. 

Our objective is twofold: (1) to develop a scalable annotation pipeline that combines LLM-based automatic labeling with targeted human verification to improve label consistency and reliability, and (2) to systematically evaluate how model scale, domain specialization, prompting strategy, and fine-tuning affect sentence-level legal argument classification performance.


This study is guided by the following research questions:  

\begin{itemize}
    \item RQ1: How effective are general-domain and legal-domain LLMs in performing sentence-level legal argument classification on U.S. caselaw? 
    \item RQ2: How can different LLM prompting strategies (zero-shot, few-shot, and chain-of-thought) be optimized to improve classification performance?
    \item RQ3: How can a scalable, high-quality legal argument mining corpus be constructed from U.S. Supreme Court and state-level caselaw using semi-automated annotation?
    
\end{itemize}


To answer these questions, we construct the LAMUS corpus through large-scale case collection, preprocessing, and sentence segmentation, followed by LLM-based automatic annotation into six categories: Fact, Issue, Rule/Law/Holding, Analysis, Conclusion, and Other, which together capture the core structure of judicial reasoning. A GPT-based verification step flags potential annotation inconsistencies, which are manually reviewed to improve label quality. We then benchmark multiple large language models, including both general-purpose and legal-domain models, under zero-shot, few-shot, and chain-of-thought prompting strategies. Performance is evaluated using accuracy, precision, recall, and F1-score, with detailed per-class analysis and fine-tuning experiments to assess robustness and generalization.


The main contributions of this work are as follows:  
(1) We introduce LAMUS, a large-scale sentence-level legal argument mining corpus for U.S. Supreme Court and Texas criminal caselaw, addressing the scarcity of annotated U.S. legal datasets.  
(2) We propose a data-centric construction pipeline that integrates LLM-based automatic annotation with targeted human verification, improving label quality and demonstrating scalable corpus development for complex legal domains.  
(3) We provide a systematic empirical evaluation of general-domain and legal-domain LLMs under multiple prompting strategies, revealing how model scale and reasoning prompts affect performance.  
(4) We demonstrate that fine-tuning substantially outperforms prompting alone for sentence-level legal argument classification, offering practical guidance for deploying LLMs in high-stakes legal NLP tasks.

\section{Related Work}

\subsection{Legal Argument Mining}

Legal argument mining (LAM) focuses on the automatic detection, classification, and structuring of argumentative components in judicial texts to support retrieval, summarization, and legal reasoning. Argument mining focuses on identifying argumentative structures such as claims, premises, and conclusions within textual data \cite{palau2009argumentation, mochales2011argumentation}. Early work relied on feature-based classifiers, while more recent studies have adopted deep learning and transformer-based models, often using sentence- or segment-level annotation schemes. Recent surveys highlight the rapid expansion of legal natural language processing research, covering tasks such as legal judgment prediction, argument mining, and legal question answering, along with specialized datasets and domain-specific models \cite{ariai2025natural, shao2025large, chi2026legalai}.

To address the high cost of manual annotation, semiautomated pipelines combining small manually labeled seeds with data augmentation techniques such as pseudolabeling and co-training have been widely explored. However, existing high-quality corpora are largely concentrated in non-U.S. jurisdictions, including ECHR-AM and LAM:ECHR, or in Chinese datasets such as CAIL. Large, state-level U.S. caselaw corpora remain underrepresented, limiting the ability of models to capture jurisdiction-specific legal reasoning styles. 

Several toolkits have been developed to facilitate natural language processing in the legal domain, including LexNLP, which provides specialized methods for extracting structured information such as legal entities, citations, and contractual clauses from legal and regulatory texts \cite{bommarito2021lexnlp}.

Additional challenges include inconsistent annotation schemes across datasets, class imbalance for minority argument types, and limited attention to long-range argument structure and reasoning chains. These limitations motivate the construction of curated U.S. caselaw resources with standardized annotation protocols and explicit quality control mechanisms.

\subsection{LLMs-as-Judge for Domain Applications}

The emergence of Large Language Models (LLMs) has introduced the paradigm of ``LLMs-as-Judges,'' in which LLMs serve as evaluators capable of assessing natural language responses across diverse tasks \cite{li2024llms}. This framework has gained attention for its scalability, broad generalization, and interpretability, as LLMs can provide natural language explanations alongside their judgments. According to \cite{li2024llms}, the LLM-as-Judge paradigm can be analyzed through five key dimensions: functionality, methodology, applications, meta-evaluation, and limitations. Functionality addresses the rationale for employing LLM judges, highlighting their potential to standardize evaluations and reduce human workload. Methodology focuses on constructing robust evaluation pipelines using LLMs, including prompt design, scoring strategies, and the aggregation of multiple judgments. Applications span domains such as education, AI system benchmarking, legal reasoning, and content moderation, demonstrating that LLMs can act as scalable evaluators in specialized contexts. Meta-evaluation examines methods for assessing the reliability and fairness of LLM judgments, while limitations highlight challenges including bias, consistency, and sensitivity to prompt formulations.

Recent work in the legal domain further emphasizes the importance of careful evaluation design. Hu et al.~\cite{hu2026evaluation} note that LLM-based systems can improve efficiency in classification tasks and reduce case complexity, but argue that evaluation must extend beyond simple output accuracy to include factors such as factual sensitivity and procedural reasoning in order to reduce misclassification or improper issue identification.

In the context of domain-specific tasks such as legal argument mining, LLMs-as-Judges show particular promise. Their ability to interpret and reason over complex legal text enables them to assess the correctness, coherence, and relevance of argument components, potentially complementing human annotators or serving as a first-pass quality control mechanism. However, prior studies indicate that reliable deployment requires careful prompt engineering, task-specific fine-tuning, and alignment with domain knowledge \cite{chalkidis-etal-2020-legal, ji2023survey}.

Recent research has also applied the LLM-as-Judge paradigm directly to legal reasoning evaluation. Frameworks such as LeMAJ extend this concept by using large language models to assess the quality of legal arguments and reasoning in specialized legal contexts \cite{li2024llms, enguehard2025lemaj}. This line of work highlights the potential of LLM-based evaluators to support scalable and structured assessment of legal NLP systems.

\subsection{Legal Intelligence with LLMs}

Large language models (LLMs) are increasingly used in legal applications such as summarization, legal question answering, statute retrieval, judgment prediction, and document drafting \cite{shao2025large}. Their ability to perform zero-shot and few-shot learning makes them attractive for legal tasks with limited labeled data, including legal argument mining. Recent work demonstrates that combining LLMs with retrieval systems or structured legal knowledge can improve accuracy and reduce hallucination. 

Large language models have also been explored for specialized legal tasks. For instance, GPT-3 has been fine-tuned to perform legal rule classification, demonstrating that general-purpose language models can be adapted to domain-specific legal reasoning tasks \cite{liga2023fine}.

At the same time, studies have raised concerns about the reliability of LLM reasoning in law. Extended or explicit reasoning does not always improve performance, particularly for smaller or non-specialized models \cite{zhang2025thinking}. Other work highlights framing effects and selective reasoning behavior in LLM-generated legal outputs \cite{cho2025modeling}. These findings suggest that effective legal NLP systems require careful evaluation of prompting strategies, model scale, and domain adaptation \cite{chi2026legalai}.

For legal argument mining, LLMs offer clear benefits for scalable annotation, data augmentation, and explainability. However, their tendency to hallucinate argumentative links and their sensitivity to prompt design necessitate hybrid approaches that combine LLMs with domain-specific models such as LegalBERT and targeted human verification. This study follows this direction by integrating LLM-based annotation with systematic quality control and benchmarking against established legal NLP models.

\section{Methodology}

\subsection{Task Definitions}

Legal argument mining (LAM) is the process of automatically identifying, classifying, and structuring argumentative components within legal texts. In this paper, our specific task is to extract legal arguments and categorize them into argument types, with a primary focus on argument classification. 

1. Argument Component Detection – identifying spans of text that represent argumentative elements such as facts, issues, rules, analysis, and conclusions.

2. Argument Classification – assigning each detected span to a predefined category of argument type, forming the foundation for downstream reasoning tasks.

We formulate this task as a sentence-level classification problem. Each sentence in a judicial opinion is treated as an individual instance, and the model is trained to assign it to one of the following categories: fact, issue, rule/law/holding, analysis, conclusion, or other.

This fine-grained classification of legal arguments provides the foundation for higher-level reasoning tasks, including the reconstruction of legal argument structures, automated legal analysis, and decision-support systems.

\begin{table*}
\caption{Annotation scheme for legal argument \cite{chen2022construction}.}
\begin{center}
\begin{tabular}{p{2.0cm}p{4.5cm}p{3.8cm}p{6.0cm}}
\hline
\textbf{Label} & \textbf{Description} & \textbf{Example} & \textbf{Annotation Guideline} \\

\hline
Fact & Any fact that is pertinent to the case. This includes testimony, statements of record, case history, and anything else that is a fact that helps establish the foundation of the case for the court to build its analysis and judgment on. & \textit {Now you say that the premises were controlled by Report Walton on that date and you know that?} & This is a very broad category.  Anything that “sets the stage,” as it were, should be labeled as a fact. Factual sentences do not include any synthesis or reasoning (that would fall under analysis); rather, they simply state events or matters of record.  
\\
\hline
Issue & Any issue or question that the court must decide. This includes the overall issue of the case as well as any sub-issues that are raised in the case. & \textit {Appellant’s sole contention on appeal is that the evidence is insufficient to sustain the conviction.} & This tends to be a pretty narrow category. Any sentence related to a question for the court to resolve should be labeled as an issue. Note that a lot of the issue sentences could be read as a fact, as any issue is inherently part of the factual history of the court case. \\
\hline
Rule/law/holding & Any statement of or reference to a rule, law, or holding. This includes sentences referencing a rule, law, or holding and then using it to reason through some point. & \textit {Reliance is handed upon Toombs v. 21, 317 S.W. 2d 737, as authority for reversal of cases where the state’s testimony was adduced by only one witness.} & Generally speaking, any time a reference is made to a rule, law, or holding, that sentence should be given this label. These sentences could be read as a fact, since whenever a law is quoted, for example, it is a fact that the law states that quotation.  \\
\hline
Analysis & Any sentence that synthesizes information to further the court’s reasoning. This includes sentences that refer to the facts of the case and then use them to push forward towards a resolution. & \textit{Had the new Rules of Appellate Procedure been in effect during the pendency of this appeal, sanctions against the responsible attorneys would have been appropriate.} &  Analysis sentences tend to move the court through the case from the facts towards the conclusion, so there is often a logical progression stringing analysis sentences together. A sentence that references a rule, law, or holding and then analyzes it in the context of the current case should be labeled Rule/Law/Holding, not as an analysis sentence.  \\
\hline
Conclusion/ opinion/answer & Any sentence that effectively resolves an issue facing the court. & \textit{The trial court erred in submitting to the jury the issue of rape by threats.} & Conclusions tend to be short and straight to the point, either agreeing or disagreeing with some argument. As there are sub-issues, there are also sub-conclusions that conclude a specific sub-issue being discussed by the court.  \\
\hline
Others & (1) Any sentence or phrase that does not fit the other labels in terms of its content. This includes section headings and others. (2) Sentences that were not split correctly.  & \textit{Ex.1: We have carefully reviewed both of these cases.} \textit{Ex.2: 21, 317 S.W. 2d 737, as authority for reversal of cases where the state’s testimony was adduced by only one witness. } & Sentences that do not add any information to the case and thus do not fall into any of the other labels. Section headings are in this category as they are neither complete sentences nor useful when building the logic of the court case. \\
\hline
\end{tabular}
\end{center}
\label{table1:annotationscheme}
\end{table*}

Table~\ref{table1:annotationscheme} summarizes the sentence-level annotation schema, including category definitions, representative examples, and annotation guidelines.

\subsection{Framework}

\begin{figure}[ht]
    \centering
    \includegraphics[width=\columnwidth]{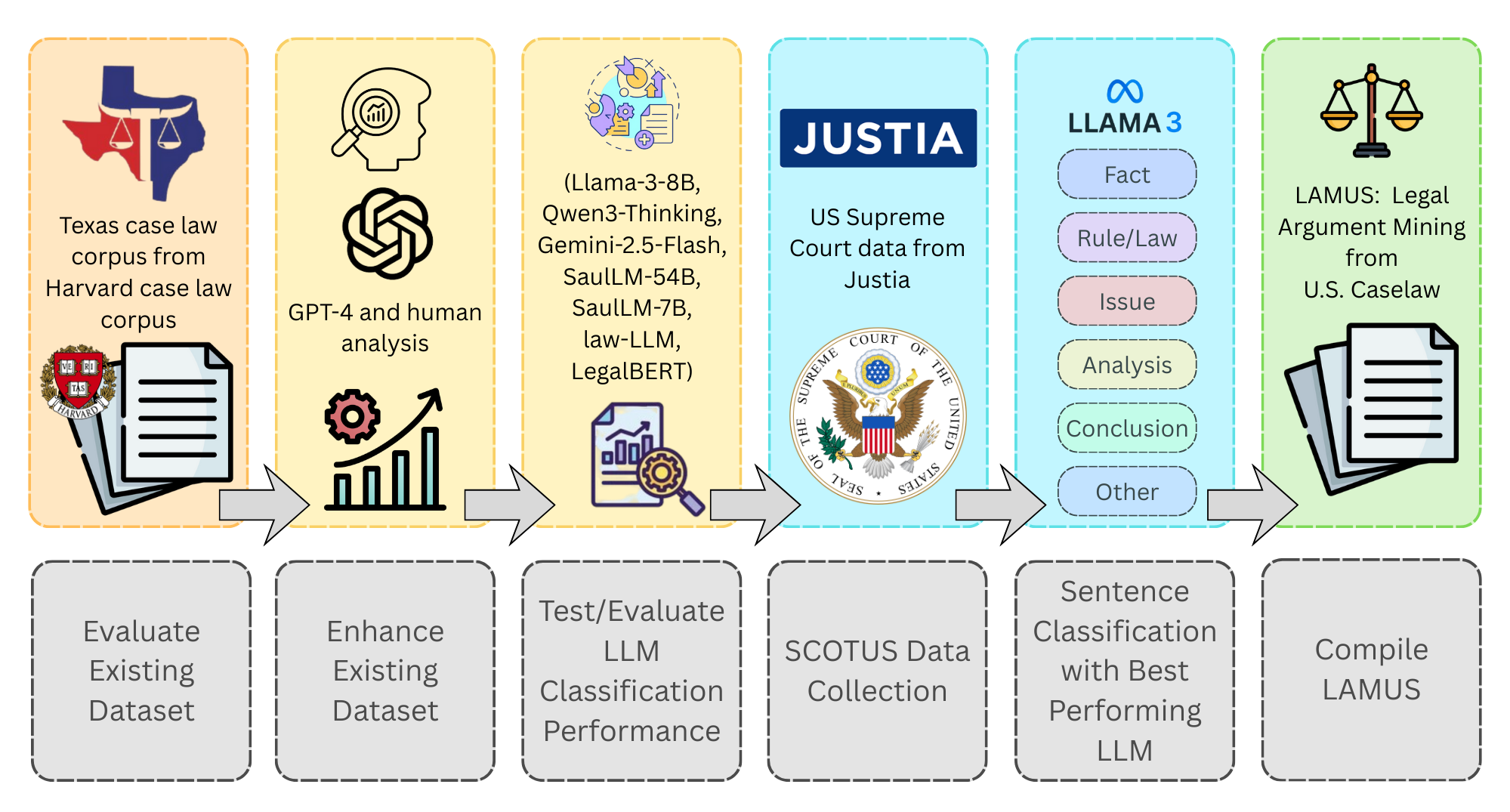}
    \caption{Methodology for corpus construction, LLM-based sentence annotation, data quality assessment, and model evaluation in the LAMUS dataset.}
    \label{fig:evolution1}
\end{figure}

Figure 1 illustrates the end-to-end workflow adopted in this study for constructing the LAMUS corpus and evaluating large language models for legal argument classification. The framework is designed as a modular, data-centric pipeline that integrates corpus acquisition, automatic annotation, quality control, and systematic model evaluation.

The workflow begins with large-scale corpus construction. Judicial opinions are collected from two primary sources: U.S. Supreme Court decisions obtained from Justia and state-level criminal caselaw from Texas. Raw case texts are cleaned, normalized, and segmented into sentences using legal-domain preprocessing tools. Invalid or noisy segments, such as section headers, citation fragments, or malformed sentences, are filtered to ensure that each unit represents a meaningful argumentative span suitable for sentence-level classification.

Next, the cleaned sentences are passed to the LLM-based automatic annotation stage. Using carefully designed prompts, large language models are instructed to assign each sentence one of the predefined legal argument categories (Fact, Issue, Rule/Law/Holding, Analysis, Conclusion, or Other). Recent research has explored methods for improving the reasoning capabilities of LLMs in legal contexts, including specialized prompting strategies and logically grounded model architectures \cite{kant2025towards}. This step enables scalable labeling across hundreds of thousands of sentences, overcoming the prohibitive cost and time requirements of fully manual annotation. Different prompt strategies (zero-shot, few-shot, and structured reasoning prompts) are employed to assess how prompting affects classification quality and model behavior.

Because automatic annotation inevitably introduces noise, the workflow incorporates an explicit data quality assessment and refinement phase. LLM predictions are compared against existing human annotations for the Texas subset, and sentences with label disagreement or high-confidence alternative predictions are flagged. These cases are then manually reviewed in context, allowing targeted correction of mislabeled instances without re-annotating the entire corpus. This hybrid verification strategy improves label consistency while maintaining scalability and reproducibility \cite{gilardi2023chatgpt}.

Following data refinement, the enhanced dataset is used for model evaluation and benchmarking. Multiple large language models with varying sizes and degrees of legal specialization are evaluated on the sentence-level classification task using standard metrics such as accuracy, precision, recall, and F1-score, reflecting growing efforts to benchmark LLM performance across legal reasoning and classification tasks \cite{guha2023legalbench, li2025legalagentbench, chi2026legalai} as well as the development of domain-adapted legal language models \cite{shu2024lawllm}.

In addition to aggregate performance, label-level behavior and distributional patterns are analyzed to assess whether models preserve the logical flow of judicial reasoning.

The final output of this pipeline is the LAMUS corpus, a structured dataset of sentence-level legal arguments with verified annotations suitable for training and evaluating legal NLP models.

Overall, this workflow provides a unified framework that connects corpus construction, LLM-based annotation, quality control, and empirical evaluation. By combining automation with targeted human oversight, the framework balances scalability and reliability, making it suitable for building large legal argument mining resources and for systematically studying the strengths and limitations of LLMs in legal NLP tasks.

\subsection{Data Quality Evaluation and Improvement}

Because the Texas criminal case law dataset was annotated by multiple human annotators, some degree of label inconsistency was unavoidable. Assigning labels such as Fact, Issue, Rule/Law/Holding, Analysis, and Other is inherently challenging in legal texts, as individual sentences may plausibly belong to more than one category. Differences in annotator interpretation therefore introduce noise and raise concerns about annotation quality. High-quality data is essential for training reliable machine learning models, as annotation errors can propagate through model predictions and degrade downstream performance \cite{chen2021data, zha2025data}.

To address this issue, we incorporated a GPT-based verification step to identify potentially mislabeled sentences. Sentences were flagged for review when the model’s predicted label disagreed with the original annotation or when the model’s confidence in an alternative label exceeded a predefined threshold. For each flagged instance, we manually re-examined the sentence in its full legal context and reassigned the label when appropriate. This hybrid approach—combining automated pre-screening with targeted human review—enabled efficient quality control without the need to re-annotate the dataset from scratch.

As a result, the overall quality of the dataset was substantially improved, with mislabeled sentences reduced by nearly one-fifth. This refinement enhances the reliability of downstream training and evaluation for legal argument classification, reduces annotation noise in subsequent experiments, and provides a stronger foundation for scaling argument mining methods to larger U.S. case law collections \cite{chen2021data}.

GPT-4 was used with the following prompt to flag potential misclassified data.

\begin{tcolorbox}[colback=gray!5, colframe=black, boxrule=1pt, rounded corners, 
    title=\textbf{Prompt for verifying legal annotations}, fonttitle=\bfseries]

\textbf{System Prompt}\\
You are a legal expert. Your task is to verify sentence annotations in U.S. legal cases. Return your response in JSON format with the following structure: \\
\vspace{2pt}
\begin{tabularx}{\linewidth}{@{}lX@{}}
\textit{Sentence:} & \textless \textit{The sentence from the case}\textgreater \\
\textit{Assigned Label:} & \textless \textit{The label currently assigned}\textgreater \\
\textit{Correct?:} & \textless \textit{YES or NO}\textgreater \\
\textit{Explanation:} & \textless \textit{If NO, brief explanation and the correct label}\textgreater \\
\end{tabularx}

\vspace{4pt}
\textbf{Annotation Categories:} \\
\begin{tabularx}{\linewidth}{@{}lX@{}}
\textbf{Fact} & Pertinent events or records establishing the case’s foundation; no reasoning. \\
\textbf{Issue} & Questions the court must decide; legal points derived from facts. \\
\textbf{Rule/Law/Holding} & References to laws, rules, or prior case holdings used to support reasoning. \\
\textbf{Analysis} & Reasoning synthesizing facts and law to progress toward a decision. \\
\textbf{Conclusion} & Resolves issues or states the court’s final decision. \\
\textbf{Others} & Non-informative text or headings not fitting above categories. \\
\end{tabularx}

\vspace{4pt}
\textbf{User Prompt} \\
Here is the sentence to verify: \textit{<Sentence from the case>} with the assigned label: \textit{<Assigned Label>}

\end{tcolorbox}

\subsection{Large Language Models}

We evaluate seven large language models (LLMs) that vary in scale, architecture, and degree of legal-domain specialization to assess their effectiveness for sentence-level legal argument classification \cite{el2024factuality}.

\begin{table}[h]
\caption{Large Language Models used in Experiment}
\label{table:modellist}
\centering
\small
\begin{tabular}{p{2.3cm}p{2.1cm}p{1.2cm}p{1.6cm}}
\hline
\textbf{Model} & \textbf{Type} & \textbf{Params} & \textbf{Domain} \\
\hline
LLaMA-3-8B & Decoder-only & 8B & General \\
\hline
Qwen3-Thinking & Decoder-only & 7B & General \\
\hline
Gemini-2.5-Flash & API-based & -- & General \\
\hline
SaulLM-54B & Legal fine-tuned & 54B & Legal \\
\hline
SaulLM-7B & Legal fine-tuned & 7B & Legal \\
\hline
law-LLM & Legal fine-tuned & 7B & Legal \\
\hline
LegalBERT & BERT Encoder & 110M & Legal \\
\hline
\end{tabular}
\end{table}

\subsubsection{General Domain LLMs}

General-domain large language models (LLMs) are trained on broad, heterogeneous corpora spanning multiple domains, enabling strong general-purpose language understanding and reasoning, but without explicit domain specialization \cite{guo2025specialized}. Prior work shows that such models can transfer effectively to specialized tasks through prompting, particularly when tasks emphasize general reasoning rather than domain-specific terminology \cite{bender2021dangers,ouyang2022training}. To evaluate their suitability for legal argument classification, we experiment with the following general-domain LLMs.

LLaMA-3-8B is an open-weight model developed by Meta that balances efficiency and reasoning capability, making it suitable for large-scale preprocessing and exploratory annotation tasks \cite{llama3}. In this study, it is used to assess the feasibility of smaller open models for sentence-level legal argument classification.

Qwen3-Thinking is a reasoning-oriented LLM designed to support structured, multi-step inference, which is particularly relevant for tasks requiring contextual interpretation and logical progression, such as legal argument mining \cite{qwen3}.

Gemini-2.5-Flash is a low-latency, high-throughput LLM developed by Google, optimized for fast inference and instruction following \cite{gemini}. Although not legally specialized, its efficiency makes it suitable for large-scale automatic annotation pipelines.

\subsubsection{Legal Domain LLMs}

Domain-specific LLMs are trained or fine-tuned on field-specific corpora, enabling them to capture specialized vocabulary, discourse structures, and reasoning patterns that are underrepresented in general-domain data \cite{guo2025specialized}. In the legal domain, exposure to statutes and case law has been shown to improve performance on legal NLP tasks requiring fine-grained argumentative distinctions \cite{chalkidis-etal-2020-legal}. The legal-domain models evaluated in this study are described below.

SaulLM-54B is a large-scale legal language model trained on extensive legal corpora, including statutes and judicial opinions, and is designed to capture complex legal reasoning patterns \cite{colombo2024saullm}. It serves as a high-capacity domain-adapted model for sentence-level legal argument classification.

Law-LLM represents a class of legal-domain LLMs optimized through pretraining or fine-tuning on legal texts \cite{chalkidis-etal-2020-legal}. In this study, it is evaluated for its ability to distinguish closely related argumentative categories.

SaulLM-7B is a smaller, computationally efficient variant of SaulLM that retains legal-domain specialization while reducing model size, allowing assessment of the trade-off between capacity and domain adaptation \cite{colombo2024saullm}.

LegalBERT is a transformer-based model adapted from BERT and further pretrained on large-scale legal corpora to capture domain-specific language patterns \cite{chalkidis-etal-2020-legal}. It serves as a supervised baseline for sentence-level legal argument classification, providing a point of comparison for evaluating LLM performance.

\subsection{Prompt Design}

The zero-shot prompt evaluates a model’s intrinsic legal reasoning by providing only task instructions and label definitions, without any task-specific examples. The model relies entirely on pretraining to map legal sentences to categories such as Fact, Issue, Rule/Law/Holding, Analysis, Conclusion, or Other. Zero-shot prompting is widely used to assess generalization and latent task competence in LLMs \cite{ouyang2022training,radford2019language,hendrycks2020measuring}, and in legal NLP, it helps determine whether a model has internalized abstract legal structures without supervision.

The few-shot prompt extends zero-shot prompting by including labeled examples for each legal argument category, providing explicit demonstrations of desired input–output behavior. This in-context learning approach allows the model to infer task structure and label semantics from a small number of exemplars \cite{ouyang2022training,min2022rethinking,liu2023pre}. In legal text classification, few-shot examples are especially useful because argument categories can be subtle and context-dependent, grounding abstract label definitions in concrete sentences to better align model predictions with established annotation schemes.

The structured chain-of-thought prompt builds on few-shot prompting by requiring explicit textual justifications and confidence scores for each classification. This encourages the model to articulate intermediate reasoning steps linking linguistic cues to the assigned label. Chain-of-thought prompting has been shown to improve performance on complex reasoning tasks by decomposing decisions into interpretable steps \cite{wei2022chain,kojima2022large,zhou2023least}. In our setting, justifications are concise and text-grounded, enhancing transparency and auditability without revealing internal model deliberations.

\begin{tcolorbox}[colback=blue!5, colframe=blue!75!black, boxrule=1pt, rounded corners, 
    title=\textbf{Zero-Shot Legal Classification Prompt}, fonttitle=\bfseries]

\textbf{System Prompt}\\
You are a legal reasoning expert. Your task is to read a legal sentence and classify each sentence into one of the following categories: Fact, Issue, Rule/Law/Holding, Analysis, Conclusion, Other. Output a list of sentences with their assigned label. 

\textbf{Annotation Categories:} \\
\begin{tabularx}{\linewidth}{@{}lX@{}}
\textbf{Fact} & Objective background details, procedural history, or evidence. \\
\textbf{Issue} & The legal question(s) the court must resolve. \\
\textbf{Rule/Law/Holding} & Stated legal rules, statutes, or holdings from precedent. \\
\textbf{Analysis} & Application of law to fact; legal reasoning or interpretation. \\
\textbf{Conclusion} & Final decisions or determinations made by the court. \\
\textbf{Other} & Sentences that do not fit the above categories. \\
\end{tabularx}

\vspace{4pt}
\textbf{User Prompt} \\
Here is the paragraph to classify: \textit{<Input Paragraph>}

\end{tcolorbox}

\begin{tcolorbox}[colback=blue!5, colframe=blue!75!black, boxrule=1pt, rounded corners, 
    title=\textbf{Few-Shot Legal Classification Prompt}, fonttitle=\bfseries]

\textbf{System Prompt}\\
You are a legal reasoning expert. Your task is to read a legal paragraph and classify each sentence into one of the following categories: Fact, Issue, Rule/Law/Holding, Analysis, Conclusion, Other. Follow this format:
\begin{tabularx}{\linewidth}{@{}lX@{}}
Sentence: & <text> \\
Label: & <Fact | Issue | Rule/Law/Holding | Analysis | Conclusion | Other> \\
\end{tabularx}

\textbf{Annotation Categories:} \\
\begin{tabularx}{\linewidth}{@{}lX@{}}
\textbf{Fact} & Background details, case history, or objective evidence. \\
\textbf{Issue} & The legal question(s) the court must answer. \\
\textbf{Rule/Law/Holding} & Stated legal rules, statutes, or holdings from precedent. \\
\textbf{Analysis} & How the law applies to the facts; legal reasoning or interpretation. \\
\textbf{Conclusion} & The final decision or holding by the court. \\
\textbf{Other} & Anything that doesn't fall into the above categories. \\
\end{tabularx}

\vspace{4pt}
\textbf{User Prompt} \\
Here is the paragraph to classify: \textit{<Input Paragraph>}

\end{tcolorbox}

\begin{tcolorbox}[colback=blue!5, colframe=blue!75!black, boxrule=1pt, rounded corners, 
    title=\textbf{Chain-of-Thought Legal Classification Prompt}, fonttitle=\bfseries]

\textbf{System Prompt}\\
You are a legal reasoning expert. For each sentence in the paragraph: 
\begin{enumerate}[leftmargin=*, itemsep=0.5em]
    \item Assign exactly one label from: Fact, Issue, Rule/Law/Holding, Analysis, Conclusion, Other.
    \item Provide a short, explicit justification (1–3 numbered points) tied to textual cues.
    \item Provide a confidence score (0–100\%) and a one-line reason for that confidence.
\end{enumerate}

\textbf{Output Format:}
\begin{tabularx}{\linewidth}{@{}lX@{}}
Sentence: & <text> \\
Label: & <one of the six> \\
Justification: & 1. <explicit cue>\\
& 2. "<how it maps to label>" \\
Confidence: & <\%> — <explanation> \\
\end{tabularx}

\vspace{4pt}
\textbf{User Prompt} \\
Here is the paragraph to classify: \textit{<Input Paragraph>}

\end{tcolorbox}

\subsection{Evaluation Metrics}

Model performance is evaluated using standard classification metrics commonly adopted in legal NLP. \textbf{Accuracy} measures the overall correctness of predictions across all classes. \textbf{Precision} quantifies the proportion of correctly predicted positive instances among all predicted positives. \textbf{Recall} (or sensitivity) measures the proportion of correctly predicted positive instances among all actual positives. \textbf{F1-score} is the harmonic mean of precision and recall.

\section{Experiment and Results}

\subsection{Data Collection}

\subsubsection{Initial Dataset}

The initial dataset for this study is derived from the Texas criminal case law corpus introduced by Chen et al. (2022), constructed from the Harvard Law Library Case Law Corpus and previously used in our work \cite{chen2022construction}. This publicly available corpus spans approximately 360 years of U.S. judicial decisions across 582 reporters. From it, all published Texas criminal cases from 1840 to 2018 were extracted, yielding 27,712 cases stored in structured JSON format, with sentence-level data used as the unit of annotation.

Full texts were segmented into sentences using LexNLP, and invalid sentences—including headings, fragments, and segmentation errors—were filtered via a supervised binary classifier trained on manually labeled examples. After preprocessing, 542,763 validated sentences were obtained, from which a subset of 5,066 sentences was manually annotated as a carefully curated seed corpus for legal argument classification \footnote{\url{https://github.com/haihua0913/legalArgumentmining}}. Annotation followed a six-class scheme grounded in the FIRAC framework: Fact, Issue, Rule/Law/Holding, Analysis, Conclusion/Opinion, and Other. Nine trained annotators (six undergraduate, three graduate) independently labeled each sentence using Doccano, with final labels determined by majority vote and adjudication when necessary. Inter-annotator agreement was measured using Cohen’s kappa, and both high- and low-agreement annotations were retained, consistent with findings that moderate disagreement does not significantly impair downstream model performance \cite{chen2022construction}.

This dataset was selected for two main reasons. First, its domain specificity ensures that argumentative structures reflect real-world judicial reasoning in U.S. criminal law, which often involves complex interactions between facts and precedent. Second, its high level of curation and quality validation makes it well suited for downstream applications such as argument classification and structure reconstruction. In this study, the Texas criminal case law dataset provides the foundation for refining annotation schemes, training and evaluating models, and benchmarking the effectiveness of large language models in legal argument mining.

\subsubsection{Data Quality Improvement Results/ Enhanced Dataset}

\begin{figure}[ht]
    \centering
    \includegraphics[width=\columnwidth]{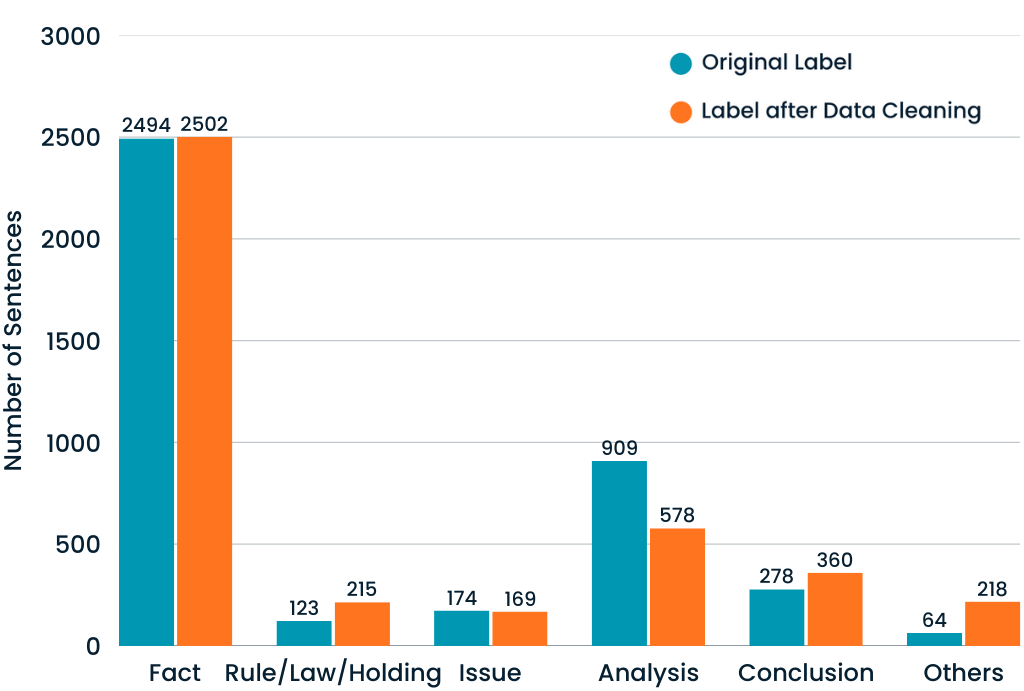}
    \caption{Comparison of Texas Case Law dataset sentence labels before and after the data cleaning process, indicating improved quality.}
    \label{fig:evolution2}
\end{figure}

To assess and improve annotation quality, we conducted a systematic verification of the annotated dataset using a GPT-based label consistency check. Each of the 4,042 annotated sentences was evaluated by comparing the original human-assigned label with GPT’s predicted label. Sentences were flagged for further inspection when the model’s predicted label differed from the original annotation or when the model expressed high confidence in an alternative classification. This automated screening step served to prioritize cases most likely to contain annotation errors.

Through this process, GPT flagged 1,058 sentences as potentially mislabeled. All flagged sentences were then manually reviewed in their full legal context by human annotators. Of these cases, 273 sentences were confirmed to be incorrectly labeled and were subsequently corrected, while the remaining flagged sentences were determined to be correctly annotated. In addition to the model-flagged cases, targeted manual spot checks were conducted on unflagged portions of the dataset to estimate residual annotation noise.

Combining the confirmed corrections from the flagged set with errors identified during additional manual review, we estimate that approximately 785 sentences in total, corresponding to 19.4\% of the dataset, carried incorrect labels prior to refinement. These findings highlight the prevalence of subtle annotation inconsistencies in sentence-level legal argument labeling and underscore the necessity of post-annotation quality control.

Following correction, the refined dataset exhibits substantially improved label consistency and reliability. Figure 2 illustrates the distributional comparison between the original dataset labels and the enhanced dataset labels after quality improvement, highlighting how the refinement process altered class proportions and reduced systematic mislabeling. The resulting enhanced dataset provides a reliable benchmark for training and evaluating legal argument classification models before applying them to large-scale corpus construction.

The final enhanced Texas dataset consists of 4,042 sentence-level annotations spanning six legal argument categories. After quality refinement, approximately 19.4\% of labels were corrected or re-evaluated, substantially reducing annotation noise and improving inter-label consistency. 

Importantly, the distribution of argument categories shifted following correction, indicating that certain classes—particularly Rule/Law/Holding and Analysis—were more prone to misclassification in the original annotations. This refined distribution better reflects the structural composition of judicial reasoning and provides a more reliable foundation for downstream model training and benchmarking.

\subsubsection{Final LAMUS Corpus}

The final LAMUS corpus consists exclusively of U.S. Supreme Court (SCOTUS) opinions automatically labeled for sentence-level legal argument structure. The corpus was constructed using a large language model (LLaMA-3-70B) that was selected based on its strong performance in the preceding classification experiments.

The refined Texas criminal case dataset was used as a high-quality benchmark and training resource for evaluating legal argument classification models. After identifying the most effective model configuration, we applied the selected model to automatically label sentences from the SCOTUS opinion corpus.

Following sentence segmentation and preprocessing, the resulting LAMUS corpus contains 2,900,083 labeled sentences spanning Supreme Court decisions from 1921 through 2025. Each sentence is annotated with one of six argument categories: Fact, Issue, Rule/Law/Holding, Analysis, Conclusion, and Other.

This large-scale corpus provides a comprehensive resource for studying judicial reasoning and supports downstream tasks such as legal argument mining, argument reconstruction, and computational analysis of Supreme Court opinions.

\subsection{Experiment Design and Settings}

To evaluate the effectiveness of large language models (LLMs) for sentence-level legal argument classification, we designed a series of experiments that systematically varied model prompting strategies, fine-tuning, and model scale.

\textbf{LLM with Different Prompting Strategies:}  
We tested each model under three distinct prompting conditions: zero-shot, few-shot, and chain-of-thought (CoT). In the zero-shot setting, models were asked to label sentences without any examples. Few-shot prompting provided labeled sentences ranging from 1 to 100 examples, systematically testing 1, 3, 4, 5, 10, 20, 40, 60, 80, and 100 examples to determine the optimal count. Examples were drawn from a curated set of 100 legal excerpts organized into 20 groups of 5 sentences (one per category). CoT prompting required models to reason step by step, encouraging systematic feature identification before producing a label. This setup allowed us to investigate how prompting interacts with model scale and domain specialization.

\textbf{LLM Fine-Tuning:}  
For certain legal-domain models, we experimented with fine-tuning on 2,585 labeled sentences from our corpus using QLoRA (4-bit quantization) with systematic hyperparameter optimization across learning rates, LoRA ranks, and training epochs. This allowed us to examine whether task-specific adaptation could improve sentence-level classification accuracy beyond what is achievable with prompting alone.

\textbf{LLM with Different Parameter Scales:}  
To explore the impact of model capacity, we included LLMs ranging from 2.5 billion to 54 billion parameters. By comparing performance across scales, we assessed the extent to which larger models benefit from multi-step reasoning (CoT) and how smaller models may be limited in representational depth when performing complex classification tasks.

\subsection{Evaluation Results}

Our experiments yield three primary findings regarding sentence-level legal argument classification. First, model performance varies substantially across model scale and domain specialization, with larger models benefiting the most from structured reasoning prompts such as Chain-of-Thought (CoT). Second, few-shot prompting consistently fails to improve performance and frequently reduces accuracy compared to zero-shot baselines, particularly for general-purpose models. Third, supervised fine-tuning provides the largest performance gains, achieving a maximum accuracy of 85.32\% with LLaMA-3-8B. These results demonstrate that while prompting can enable competitive performance, task-specific fine-tuning remains the most effective approach for high-accuracy legal argument classification.

\subsubsection{Overall Results}

Table~\ref{table:llm_results} summarizes classification accuracy across models and prompting strategies. The highest performance among prompted models was achieved by LLaMA-3-8B with Chain-of-Thought prompting (75.89\%), followed by SaulLM-54B with CoT (72.80\%). Zero-shot prompting yielded competitive results for domain-specialized models, with SaulLM-54B reaching 67.39\% accuracy. In contrast, few-shot prompting consistently underperformed zero-shot baselines across evaluated models, indicating that in-context demonstrations do not reliably enhance legal sentence classification.

The results, summarized in Table~\ref{table:llm_results}, show that model performance varies with both model capacity and domain specialization. General-purpose models, such as LLaMA-3-8B, achieve the highest performance when combined with Chain-of-Thought (CoT) prompting, reaching 75.89\% accuracy. Smaller legal-specialized models, such as SaulLM-7B, perform better under zero-shot conditions, likely due to limited representational capacity for multi-step reasoning. The largest legal-domain model, SaulLM-54B, benefits from CoT prompting as well, achieving 72.80\% accuracy, slightly below its zero-shot performance of 67.39\%, indicating that domain expertise can partially compensate for smaller-scale reasoning. Other models, including law-LLM and Qwen3-Thinking, performed best with zero-shot prompting, while Gemini-2.5-Flash performed near random (5.41\%) across all conditions, likely due to limited model capacity and parsing issues.

Overall, these results demonstrate that large-scale models combined with CoT prompting can substantially improve sentence-level legal argument classification, emphasizing the importance of aligning prompting strategy with both model capacity and domain specialization.

\begin{table}
\begin{threeparttable}
\caption{Zero-Shot, Few-Shot, and Chain-of-Thought (CoT) performance of evaluated models. Bold indicates the best performance. Stability verified through 10 independent runs (mean 74.71\% ± 0.56\%, p < 0.001).}
\centering
\begin{tabular}{lcccc}
\hline
\textbf{Model} & \textbf{Domain} & \textbf{Zero-Shot} & \textbf{Few-Shot} & \textbf{CoT} \\
\hline
LLaMA-3-8B & General & 65.38\% & 45.75\% & \textbf{75.89\%}  \\
\hline
SaulLM-54B & Legal & 67.39\% & 64.76\% & \textbf{72.80\%}  \\
\hline
law-LLM & Legal & \textbf{60.12\%} & 31.68\% & 28.75\%  \\
\hline
Qwen3-Thinking & General & \textbf{56.11\%} & 49.30\% & 54.10\%  \\
\hline
SaulLM-7B & Legal & \textbf{52.09\%} & 21.64\% & 38.02\%  \\
\hline
Gemini-2.5-Flash & General & 5.41\% & 5.41\% & 5.41\% \\
\hline
\end{tabular}
\label{table:llm_results}
\end{threeparttable}
\end{table}

\subsubsection{Performance of different Prompting Strategies}

Among the evaluated models, the effectiveness of prompting strategies varies significantly. CoT prompting outperformed zero-shot and few-shot approaches for large, general-purpose models, such as LLaMA-3-8B, indicating that explicit stepwise reasoning helps models systematically identify relevant sentence features. Smaller models, including SaulLM-7B, were better suited to zero-shot prompting due to limited reasoning capacity. These observations highlight that CoT is not universally beneficial and interacts strongly with both model scale and domain expertise.

\subsubsection{The Impact of Few-Shot Samples}

Few-shot prompting consistently reduced performance for the models fully evaluated in this study. Table~\ref{table:fewshot_full} reports results for LLaMA-3-8B and SaulLM-54B, the two models for which all prompting configurations were completed. 

For LLaMA-3-8B, few-shot prompting resulted in a substantial degradation in accuracy relative to the zero-shot baseline, with decreases ranging from 12\% to 17\% depending on the number of examples provided. Notably, increasing the number of examples did not recover zero-shot performance, suggesting that few-shot demonstrations introduce noise or encourage overfitting to prompt structure rather than improving task understanding.

SaulLM-54B exhibited more stable behavior, with accuracy varying across few-shot settings but remaining broadly comparable to the zero-shot baseline. This contrast indicates that the negative impact of few-shot prompting is model-dependent and particularly pronounced for general-purpose LLMs in legal sentence classification tasks.

Due to GPU availability constraints, additional models (SaulLM-7B, law-LLM, and Qwen3-Thinking) could not be fully evaluated. Nevertheless, the completed experiments are sufficient to establish the key finding that few-shot prompting does not reliably improve—and may significantly harm—performance in this setting.

\textbf{Extended Few-Shot Analysis (5--100 Examples).} To comprehensively investigate the relationship between example count and classification accuracy, we extended our analysis to test 5, 10, 20, 40, 60, 80, and 100 examples using LLaMA-3-8B (Table~\ref{table:fewshot_extended}). The results reveal a consistent pattern of performance degradation: accuracy decreased from 67.23\% (zero-shot baseline) to 65.07\% with 5 examples, and continued declining monotonically to 53.94\% with 100 examples---a 13.29 percentage point decrease from the baseline.

This negative result indicates that for general-purpose models on domain-specific legal classification tasks, fixed few-shot examples introduce noise rather than useful guidance. We hypothesize this stems from domain mismatch: the curated examples represent generic legal scenarios, while the test data consists of specific Texas criminal court language with distinct terminology and sentence structure. This finding has important implications for legal NLP practitioners: generic few-shot examples may harm rather than help performance on jurisdiction-specific classification tasks.

\begin{table}[h]
\caption{Impact of few-shot prompting on model performance across completed experiments.}
\begin{center}
\begin{tabular}{p{1.8cm}p{0.9cm}p{0.9cm}p{0.9cm}p{0.9cm}p{0.9cm}}
\hline
\textbf{Model} & \textbf{Zero-Shot} & \textbf{1-ex} & \textbf{3-ex} & \textbf{4-ex} & \textbf{5-ex} \\
\hline
LLaMA-3-8B & 65.38\% & 47.76\% & 49.61\% & 52.86\% & 50.54\% \\
\hline
SaulLM-54B & 67.39\% & 54.40\% & 66.31\% & 59.51\% & 67.70\% \\
\hline
\end{tabular}
\end{center}
\label{table:fewshot_drop}
\end{table}
\footnotetext{Zero-shot baseline varies slightly between experiments (65.38\% vs 67.23\%) due to differences in prompt formatting. Table~\ref{table:fewshot_drop} uses the original prompt template, while Table~\ref{table:fewshot_extended} uses a standardized template for fair comparison across the 5--100 example sweep.}

\begin{table}[h]
\caption{Extended few-shot analysis (5--100 examples) for LLaMA-3-8B. Performance degrades monotonically as example count increases.}
\begin{center}
\begin{tabular}{p{2.0cm}p{1.8cm}p{2.4cm}}
\hline
\textbf{\# Examples} & \textbf{Accuracy} & \textbf{$\Delta$ vs Zero-Shot} \\
\hline
0 (Zero-Shot) & 67.23\% & baseline \\
\hline
5 & 65.07\% & -2.16\% \\
\hline
10 & 66.15\% & -1.08\% \\
\hline
20 & 64.91\% & -2.32\% \\
\hline
40 & 65.53\% & -1.70\% \\
\hline
60 & 60.43\% & -6.80\% \\
\hline
80 & 59.04\% & -8.19\% \\
\hline
100 & 53.94\% & -13.29\% \\
\hline
\end{tabular}
\end{center}
\label{table:fewshot_extended}
\end{table}

\begin{figure}[ht]
    \centering
    \includegraphics[width=\columnwidth]{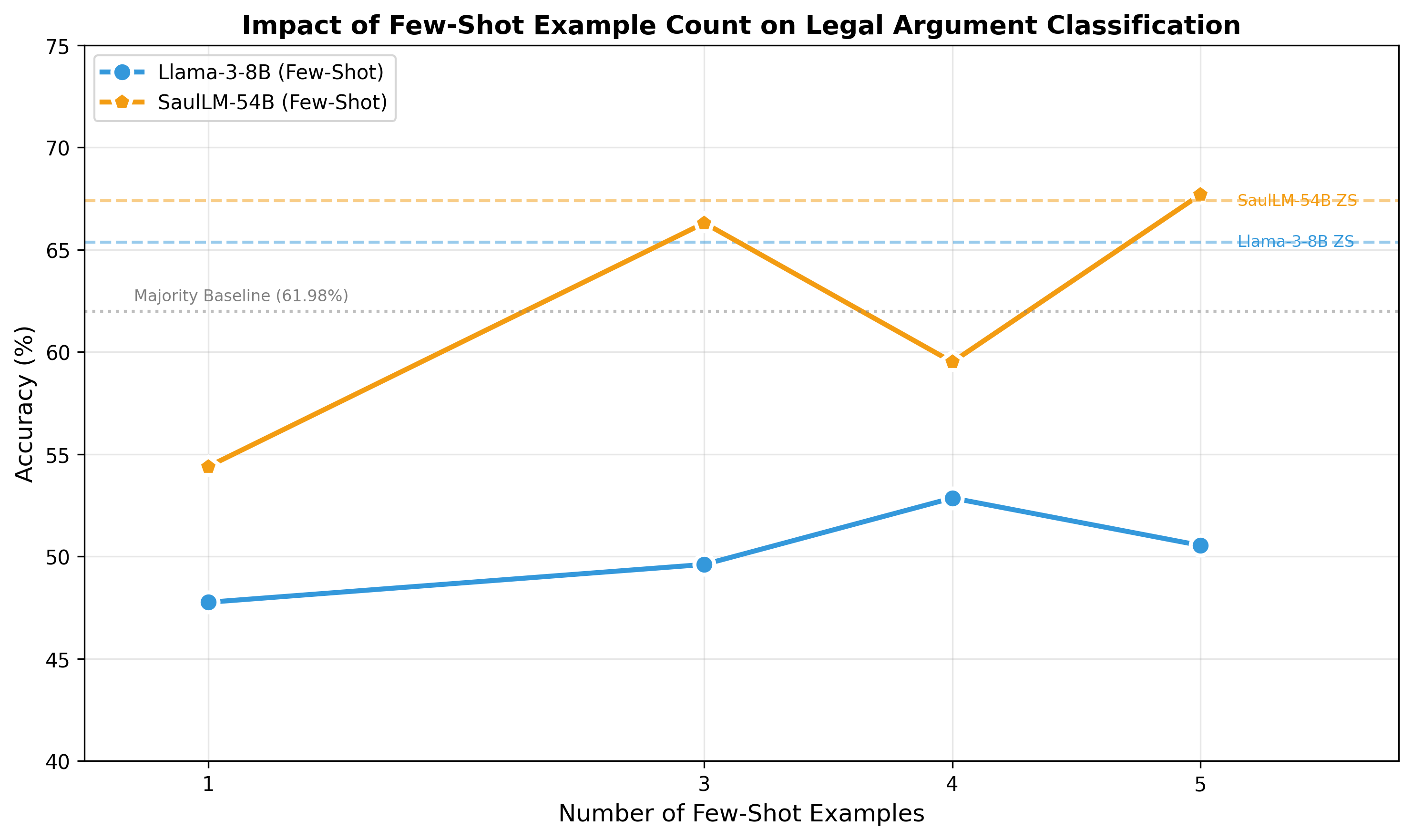}
    \caption{Accuracy of LLaMA-3-8B and SaulLM-54B under varying few-shot example counts. Few-shot prompting harms performance for LLaMA-3-8B, while SaulLM-54B remains relatively stable.}
    \label{fig:evolution3}
\end{figure}

\subsubsection{Fine Tuning}

Fine-tuning substantially improves performance over prompting strategies. LLaMA-3-8B achieved the highest accuracy of 85.32\% after fine-tuning, representing a 23.18\% improvement over the majority-class baseline. LegalBERT also performed well at 81.30\%, confirming that domain-specialized models benefit from task-specific adaptation. Even the original fine-tuned LLaMA-3-8B (without ablation) surpassed 80\%, showing that careful hyperparameter tuning can yield further gains.

Table~\ref{table:legalbert_per_class} provides per-class performance for LegalBERT. The model achieves the highest precision and recall on the “Facts” category, while performance drops on “Rule/Law/Holding” and “Others,” highlighting the challenges of classifying nuanced or infrequent legal arguments.

\begin{table}[h]
\caption{Accuracy of fine-tuned models and improvement over baseline.}
\begin{center}
\begin{tabular}{p{1.8cm}p{2.6cm}p{1.4cm}p{1.5cm}}
\hline
\textbf{Model} & \textbf{Method} & \textbf{Accuracy} & \textbf{Baseline} \\
\hline
LLaMA-3-8B & Fine-tuned (Ablation Best) & \textbf{85.32\%} & +23.34\% \\
\hline
LegalBERT & Fine-tuned & 81.30\% & +19.32\% \\
\hline
LLaMA-3-8B & Fine-tuned (Original) & 80.37\% & +18.39\% \\
\hline
\end{tabular}
\end{center}
\label{table:fine_tuned_accuracy}
\end{table}

\begin{table}[h]
\caption{Per-class performance metrics for LegalBERT on sentence-level legal argument classification. (LLaMA)}
\begin{center}
\begin{tabular}{p{2.2cm}p{1.2cm}p{1.1cm}p{1.0cm}p{1.2cm}}
\hline
\textbf{Category} & \textbf{Precision} & \textbf{Recall} & \textbf{F1} & \textbf{Support} \\
\hline
Facts & 0.91 & 0.94 & 0.92 & 401 \\
\hline
Issue & 0.86 & 0.70 & 0.78 & 27 \\
\hline
Rule/Law/Holding & 0.79 & 0.32 & 0.46 & 34 \\
\hline
Analysis & 0.56 & 0.67 & 0.61 & 92 \\
\hline
Conclusion & 0.67 & 0.76 & 0.71 & 58 \\
\hline
Others & 0.70 & 0.40 & 0.51 & 35 \\
\hline
Weighted Avg & 0.82 & 0.81 & 0.81 & 647 \\
\hline
\end{tabular}
\end{center}
\label{table:legalbert_per_class}
\end{table}

\subsubsection{Ablation Study}

The ablation study (Table~\ref{table:ablation_results}) shows the effect of hyperparameter choices on LLaMA-3-8B. The best configuration (learning rate 2e-4, 3 epochs, LoRA rank 16) achieved 85.32\% accuracy. Learning rate had the largest impact (±10\%), while epochs and LoRA rank were moderately or minimally sensitive. Table~\ref{table:hyperparam_sensitivity} summarizes the sensitivity of each hyperparameter.

\begin{table}[h]
\caption{Complete ablation study results for fine-tuned LLaMA-3-8B. The best configuration is highlighted.}
\begin{center}
\begin{tabular}{p{1.2cm}p{1.4cm}p{1.2cm}p{1.2cm}p{1.6cm}}
\hline
\textbf{Config} & \textbf{Learning Rate} & \textbf{Epochs} & \textbf{LoRA Rank} & \textbf{Accuracy} \\
\hline
best & 1e-4 & 5 & 8 & \textbf{85.32\%} \\
\hline
lr\_1e-4 & 1e-4 & 5 & 16 & 85.32\% \\
\hline
lr\_2e-4 & 2e-4 & 3 & 16 & 85.16\% \\
\hline
lora\_32 & 5e-5 & 5 & 32 & 85.01\% \\
\hline
epoch\_5 & 1e-4 & 5 & 32 & 85.01\% \\
\hline
lora\_8 & 2e-4 & 5 & 8 & 84.85\% \\
\hline
lr\_2e-4 & 2e-4 & 3 & 8 & 84.70\% \\
\hline
lora\_16 & 5e-5 & 5 & 16 & 84.70\% \\
\hline
lr\_1e-4 & 1e-4 & 3 & 16 & 84.54\% \\
\hline
lora\_32 & 2e-4 & 3 & 32 & 84.39\% \\
\hline
\end{tabular}
\end{center}
\label{table:ablation_results}
\end{table}

\begin{figure*}[t]
    \centering
    \includegraphics[width=\textwidth]{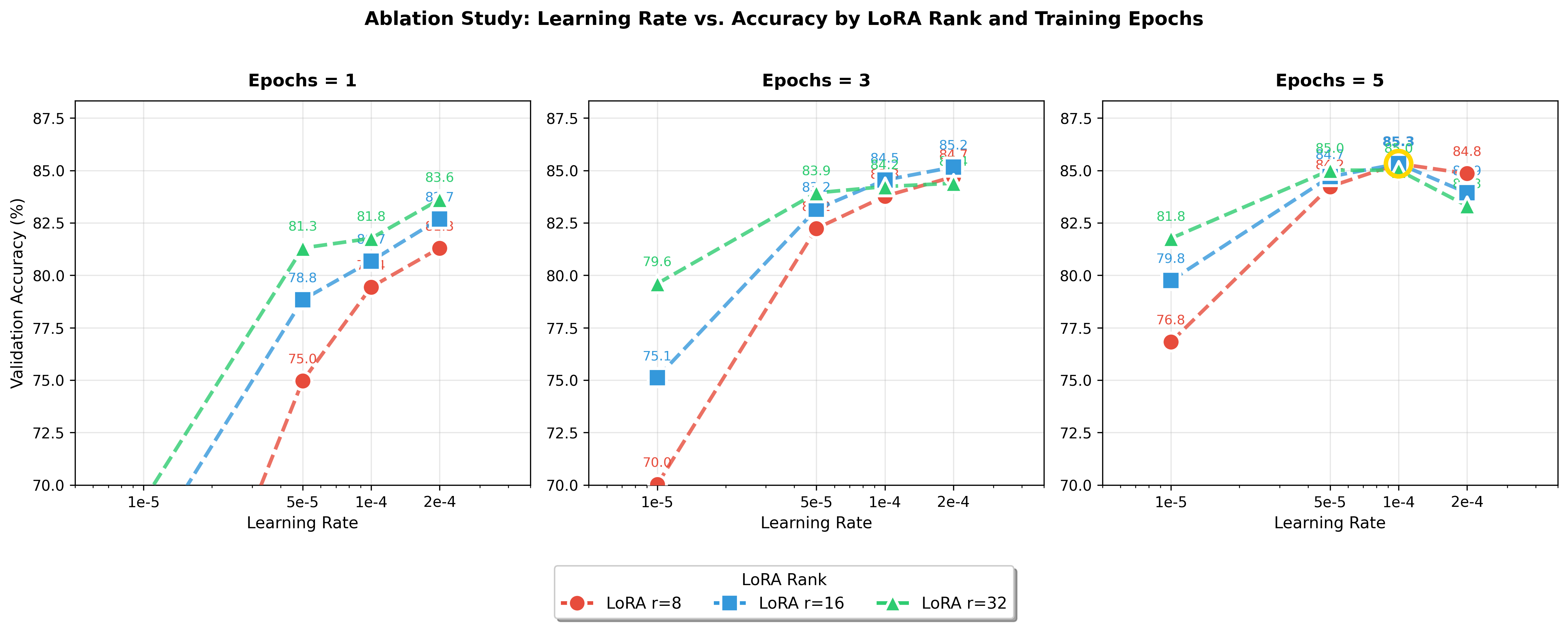}
    \caption{Ablation Study: Learning Rate vs. Accuracy by LoRA Rank and Training Epochs. Results from 36 experiments across 4 learning rates (1e-5, 5e-5, 1e-4, 2e-4), 3 LoRA ranks (8, 16, 32), and 3 epoch settings (1, 3, 5). The best configuration (LR=1e-4, Rank=8, Epochs=5) achieves 85.32\% accuracy.}
    \label{fig:ablation_study}
\end{figure*}

\begin{table}[h]
\caption{Hyperparameter sensitivity analysis for fine-tuned LLaMA-3-8B based on 36 ablation experiments.}
\begin{center}
\begin{tabular}{p{1.8cm}p{1.6cm}p{1.5cm}p{2.2cm}}
\hline
\textbf{Parameter} & \textbf{Sensitivity} & \textbf{Optimal Value} & \textbf{Impact on Accuracy} \\
\hline
Learning Rate & HIGH & 1e-4 & $\pm$10\% \\
\hline
Epochs & MODERATE & 5 & $\pm$2.5\% \\
\hline
LoRA Rank & LOW & 8 & $\pm$0.8\% \\
\hline
\end{tabular}
\end{center}
\label{table:hyperparam_sensitivity}
\end{table}


\subsection{Discussions}

\subsubsection{Why Chain-of-Thought Prompting Outperforms Few-Shot Prompting}

The observed performance gains from Chain-of-Thought (CoT) prompting can be attributed to its ability to explicitly induce step-by-step reasoning within the model. By requiring intermediate reasoning steps, CoT encourages systematic identification of legally relevant features—such as factual predicates, normative rules, and analytical conclusions—rather than allowing the model to jump directly to a label. In contrast, few-shot prompting tends to promote surface-level pattern matching, where the model aligns new inputs with previously seen examples without fully internalizing the underlying classification criteria. This distinction is particularly salient for sentence-level legal argument classification, where subtle semantic differences separate closely related categories.

Model capacity further mediates the effectiveness of CoT prompting. Our results indicate that only sufficiently capable models ($\geq$8B parameters) consistently benefit from CoT, suggesting that smaller models lack the representational depth required to maintain coherent multi-step reasoning. For these models, the added reasoning burden may introduce noise rather than clarity, diminishing any potential gains. This finding reinforces the notion that CoT is not universally beneficial but instead interacts strongly with model scale and reasoning capacity.

\subsubsection{Why General-Purpose Models Outperform Legal-Domain Models}

Despite their domain specialization, legal-specific language models underperform general-purpose models such as LLaMA-3-8B-Instruct in this task. One explanation lies in differences in training data quality and diversity. LLaMA models are trained on substantially larger and more heterogeneous corpora, enabling broader semantic coverage and more robust generalization. This diversity appears advantageous for sentence-level classification, which requires flexible interpretation of legal language across varying contexts rather than strict adherence to domain-specific templates.

Additionally, instruction-tuned general models demonstrate superior adherence to task instructions. LLaMA-3-8B-Instruct in particular, is explicitly optimized for following structured prompts, which likely enhances its responsiveness to CoT and classification directives. In contrast, legal-domain models may be overfitted to other legal NLP objectives—such as document retrieval or citation prediction—leading to reduced adaptability when confronted with fine-grained argument role classification. This over-specialization may limit their ability to generalize beyond the specific distributions encountered during pretraining.

\subsubsection{Why Fine-Tuning Produces Substantial Performance Gains}

Fine-tuning yields the largest performance improvement (+9.27\%), underscoring its effectiveness for this task. Through supervised fine-tuning, the model directly learns task-specific vocabulary, stylistic cues, and semantic patterns associated with each argumentative category. This process sharpens decision boundaries between conceptually similar labels, such as Rule and Analysis, which are particularly challenging to distinguish using prompting alone.

Moreover, fine-tuning resolves several practical limitations of prompt-based approaches. By eliminating reliance on free-form generation, it reduces output parsing errors and ensures consistent label formatting. The use of label masking further concentrates learning on the classification objective, preventing the model from expending capacity on irrelevant token prediction. Together, these factors explain the pronounced gains achieved through fine-tuning and highlight its importance for high-precision legal argument classification.

\section{LAMUS Corpus Construction using LLMs}

The construction pipeline therefore proceeds in two stages: (1) model evaluation using the manually curated Texas criminal case dataset and (2) large-scale automatic labeling of SCOTUS opinions to construct the LAMUS corpus.

\subsection{Data Sources}

For this study, we constructed the \textbf{LAMUS corpus}, a comprehensive dataset of United States Supreme Court (SCOTUS) opinions spanning from 1759 to the present. To compile the corpus, we developed a custom web scraper that systematically retrieved case texts from the \textit{Justia} legal database, which provides open access to court opinions. The dataset includes all reported cases, beginning with early Pennsylvania cases and continuing through the Roberts Court, ensuring coverage of the full historical breadth of the US Supreme Court \footnote{\url{https://supreme.justia.com/cases/federal/us/}}.

Each case in the corpus contains the full opinion text, including majority, concurring, and dissenting opinions where available. Metadata such as the case name, citation, date of decision, and participating justices were also extracted to enable structured analysis. The resulting corpus represents one of the most extensive publicly accessible compilations of Supreme Court decisions, providing a rich resource for research in legal argument mining, natural language processing, and computational legal studies.

\subsection{Data Labeling using LLMs}

To annotate the LAMUS corpus with argumentative structure, we employed the state-of-the-art large language model \textbf{LLaMA-3-70B}, which demonstrated the highest performance in our preliminary evaluations. The model was tasked with labeling each sentence in the corpus according to its argumentative role, including identification of claims, premises, and supporting or opposing reasoning. 

By leveraging LLaMA-3-70B, we were able to efficiently generate high-quality sentence-level annotations across the full dataset, capturing nuanced legal reasoning that would be challenging to label manually. This approach enables subsequent analyses such as automated argument reconstruction, argument generation, and the study of rhetorical patterns within Supreme Court opinions. 

The use of a highly capable LLM ensures that the labeled data maintains both consistency and depth, providing a reliable foundation for downstream computational legal tasks.

\subsection{Human Verification}

To validate the quality of our LLM-based automatic labeling pipeline, we conducted human verification following standard inter-annotator agreement protocols. Two expert annotators independently labeled a stratified random sample of 600 sentences from the SCOTUS labeled dataset, with 100 sentences per argument category ensuring balanced representation across all six labels.

Both annotators received identical instructions defining the six categories (Fact, Issue, Rule/Law/Holding, Analysis, Conclusion, Other) with examples consistent with the annotation guidelines in Table~\ref{table1:annotationscheme}. Each annotator completed their labels independently without discussion to ensure unbiased assessment. Inter-annotator agreement was measured using Cohen's Kappa ($\kappa$), which accounts for agreement occurring by chance.

Table~\ref{table:kappa_results} summarizes the human verification results. The annotators achieved a Cohen's Kappa of $\kappa = 0.85$, indicating almost perfect agreement according to standard interpretation scales \cite{landis1977measurement}. Direct agreement (percentage of identical labels) was 87.3\%, with only 76 sentences (12.7\%) showing disagreement between annotators, primarily in edge cases involving sentences with characteristics of multiple categories.

The high agreement between human annotators and model predictions (89.2\% average) validates the reliability of our automated labeling pipeline. Annotator 1 agreed with model predictions in 90.5\% of cases, while Annotator 2 agreed in 87.8\% of cases. These results confirm that the LAMUS corpus maintains annotation quality suitable for downstream legal NLP research and that our LLM-based labeling approach produces labels consistent with expert human judgment.

\begin{table}[h]
\caption{Human verification results for SCOTUS labeled data (N = 600 sentences).}
\begin{center}
\begin{tabular}{p{4.5cm}p{3.0cm}}
\hline
\textbf{Metric} & \textbf{Value} \\
\hline
Total Sentences Verified & 600 \\
\hline
Sentences per Category & 100 \\
\hline
Number of Annotators & 2 \\
\hline
Cohen's Kappa ($\kappa$) & 0.85 \\
\hline
Interpretation & Almost Perfect \\
\hline
Direct Agreement & 87.3\% \\
\hline
Annotator 1 vs Model & 90.5\% \\
\hline
Annotator 2 vs Model & 87.8\% \\
\hline
Average Human-Model & 89.2\% \\
\hline
Total Disagreements & 76 (12.7\%) \\
\hline
\end{tabular}
\end{center}
\label{table:kappa_results}
\end{table}

\subsection{Statistics of the Corpus}

The LAMUS corpus represents one of the largest structured resources for sentence-level legal argument mining in U.S. case law. It combines a high-quality manually annotated dataset derived from Texas criminal appellate decisions with a large-scale automatically labeled corpus of U.S. Supreme Court opinions. Together, these datasets provide both reliable ground-truth annotations for benchmarking and extensive coverage for large-scale computational analysis.

After preprocessing and sentence segmentation, the SCOTUS portion of the corpus contains 2,900,083 labeled sentences across six argument categories. These sentences span eight Supreme Court eras from the Taft Court (1921) through the Roberts Court (2025), providing substantial historical coverage for studying the evolution of legal reasoning.

\begin{figure}[ht]
    \centering
    \includegraphics[width=\columnwidth]{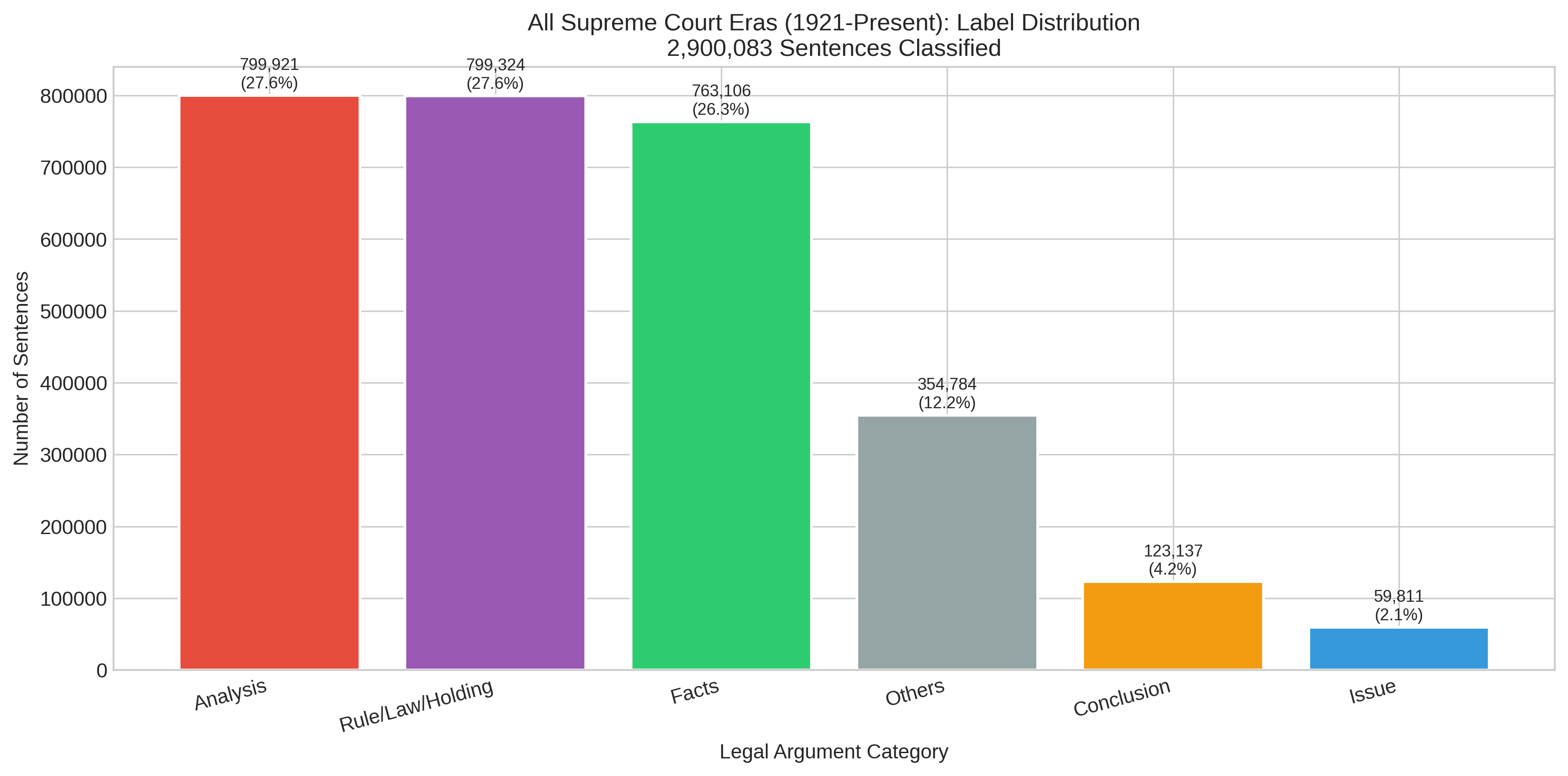}
    \caption{Overall distribution of legal argument label categories across
all Supreme Court eras (1921--present), comprising 2{,}900{,}083 sentences.
The corpus is dominated by \emph{Analysis}, \emph{Rule/Law/Holding}, and
\emph{Facts}, while \emph{Issue} and \emph{Conclusion} occur less
frequently but consistently.}
\label{fig:label-distribution}
\end{figure}

\begin{figure}[ht]
    \centering
    \includegraphics[width=\columnwidth]{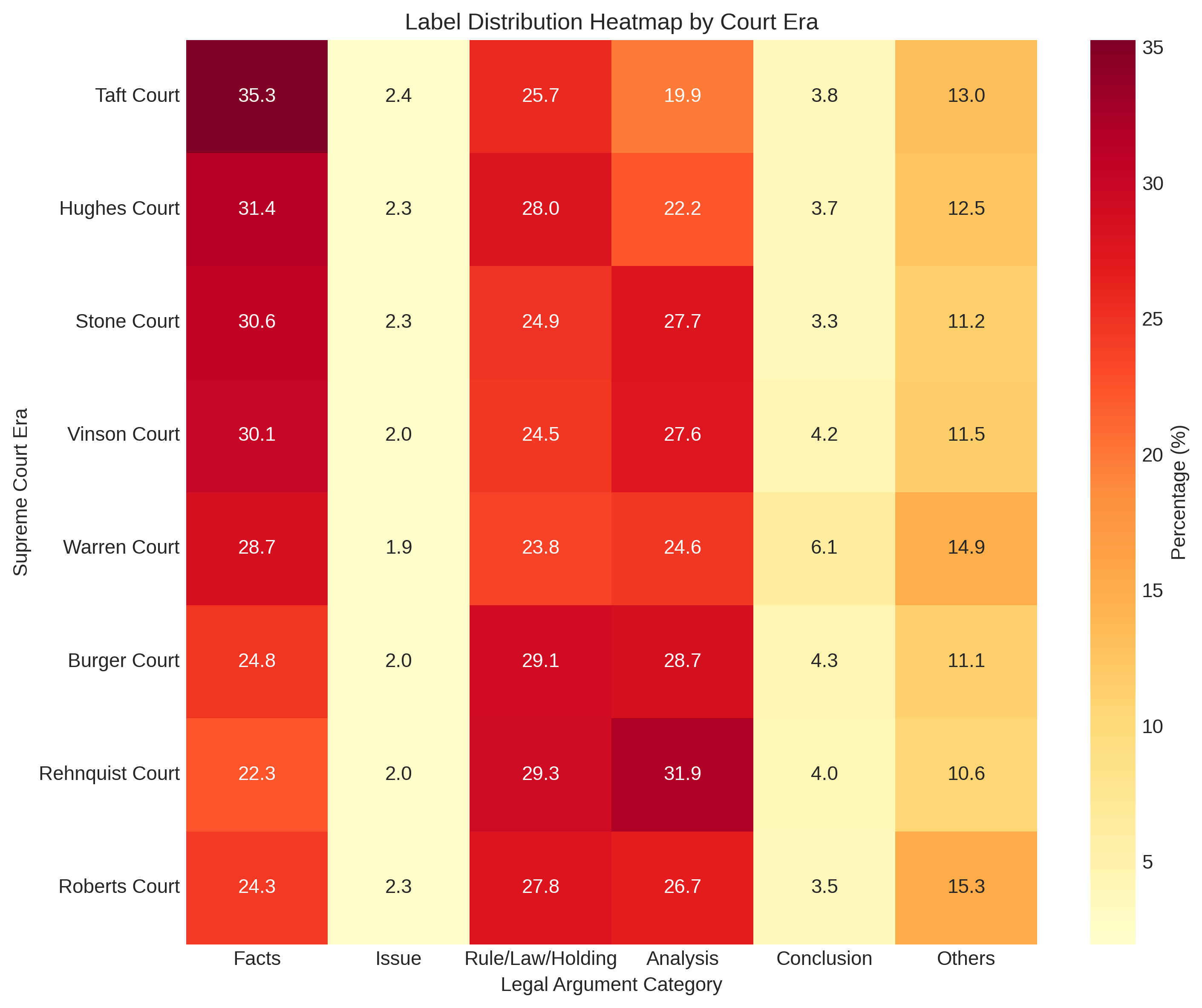}
    \caption{Distribution of legal argument label categories across Supreme
Court eras. Each cell shows the percentage of sentences assigned to a
label (\emph{Facts}, \emph{Issue}, \emph{Rule/Law/Holding}, \emph{Analysis},
\emph{Conclusion}, \emph{Others}) within a given court era. Although the
same core labels dominate across all periods, their relative proportions
vary over time.}
\label{fig:label-by-court}
\end{figure}

Figures~\ref{fig:label-by-court} and~\ref{fig:label-distribution} illustrate the distribution of legal argument label categories in the LAMUS corpus. The figures report the prevalence of six labels—\emph{Facts}, \emph{Issue}, \emph{Rule/Law/Holding}, \emph{Analysis}, \emph{Conclusion}, and \emph{Others}—both across Supreme Court eras and over the full dataset of sentences.

Table~\ref{table:corpus_stats} summarizes the distribution of sentences across Supreme Court eras. The Burger Court (1969–1986) constitutes the largest portion of the dataset, followed by the Rehnquist and Warren Courts. More recent decisions from the Roberts Court (2005–2025) account for 362,891 labeled sentences, enabling contemporary legal analysis while preserving historical coverage.

The corpus exhibits substantial temporal diversity, capturing shifts in legal reasoning and argumentative style across more than a century of jurisprudence. This broad coverage supports both diachronic analysis and robust model training, reducing bias toward any single judicial era.

All labeled data are released publicly to support reproducibility and future research. We provide two versions of the dataset: (i) a comprehensive corpus covering all court eras and (ii) a Roberts Court–only subset to facilitate focused analysis of modern jurisprudence.

\begin{table}[h]
\caption{Distribution of labeled sentences across Supreme Court eras in the LAMUS corpus.}
\begin{center}
\begin{tabular}{p{2.6cm}p{2.0cm}p{2.0cm}}
\hline
\textbf{Court Era} & \textbf{Years} & \textbf{Sentences} \\
\hline
Burger Court & 1969--1986 & 809,409 \\
Rehnquist Court & 1986--2005 & 673,564 \\
Warren Court & 1953--1969 & 377,645 \\
Roberts Court & 2005--2025 & 362,891 \\
Hughes Court & 1930--1941 & 213,122 \\
Vinson Court & 1946--1953 & 170,975 \\
Taft Court & 1921--1930 & 155,066 \\
Stone Court & 1941--1946 & 137,411 \\
\hline
\textbf{Total} & 1921--2025 & \textbf{2,900,083} \\
\hline
\end{tabular}
\end{center}
\label{table:corpus_stats}
\end{table}

\begin{figure}[ht]
    \centering
    \includegraphics[width=\columnwidth]{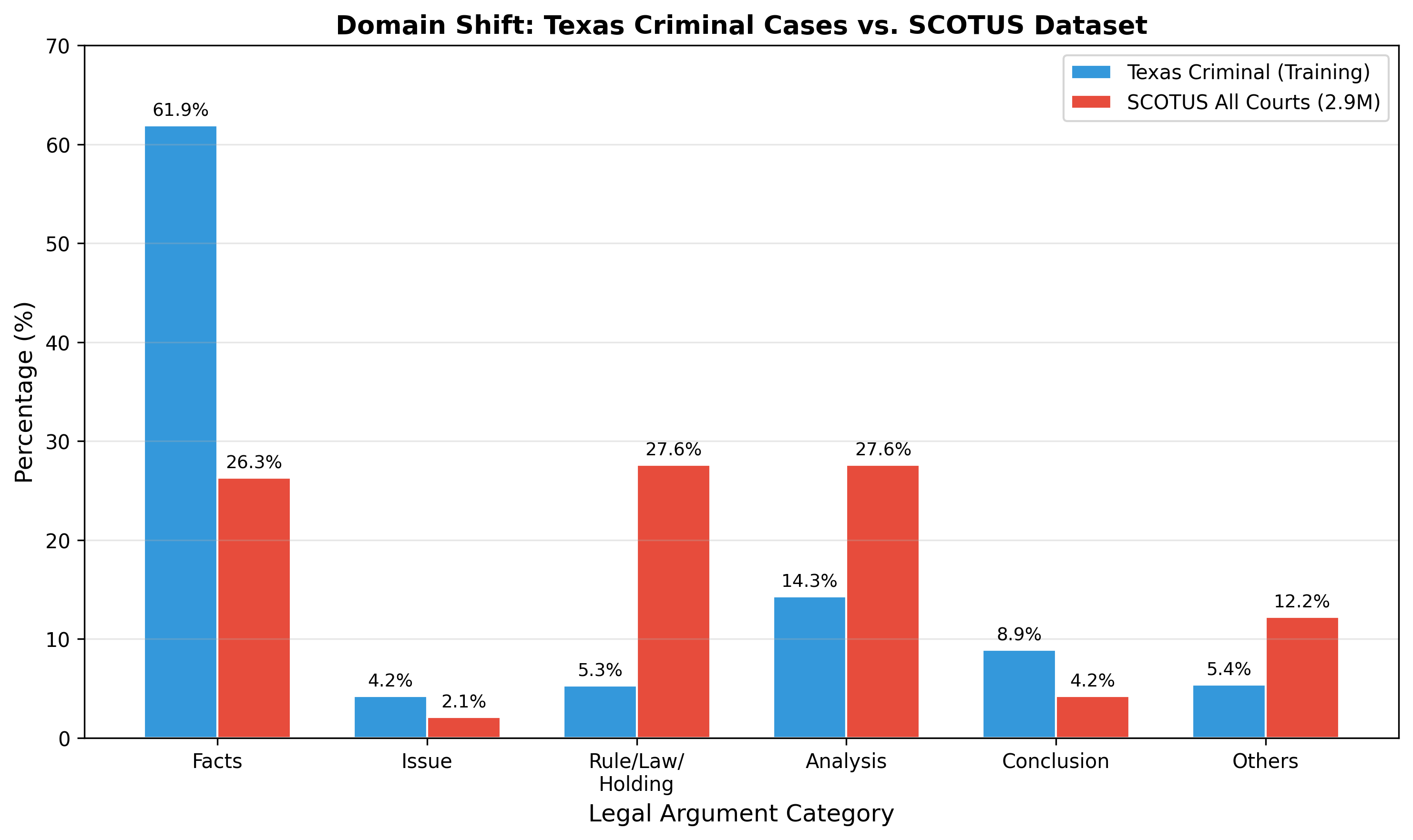}
    \caption{Comparison of legal argument label distributions in the Texas Criminal Cases training corpus and the SCOTUS dataset, highlighting substantial domain shift in argumentative structure across courts.}
\label{fig:domain_shift}
\end{figure}

Figure~\ref{fig:domain_shift} illustrates a substantial domain shift between the Texas Criminal Cases training corpus and the SCOTUS dataset. The Texas corpus is heavily dominated by \emph{Facts} (61.9\%), whereas the SCOTUS dataset exhibits a more balanced distribution, with \emph{Rule/Law/Holding} and \emph{Analysis} each accounting for 27.6\% of the data. This contrast highlights differing argumentative emphases between state-level criminal opinions and U.S. Supreme Court decisions.

\subsection{Implications}

The LAMUS corpus provides a foundational resource for a wide range of computational legal research. Its comprehensive coverage of US Supreme Court decisions enables the development and evaluation of advanced legal AI systems, particularly in the domains of \textit{argument generation} and \textit{argument reconstruction}. 

By offering full opinion texts along with structured metadata, the corpus allows models to identify and analyze the reasoning patterns employed by the Court over time. Researchers can leverage it to automatically extract legal arguments, reconstruct chains of reasoning, and generate novel arguments grounded in historical precedent. Furthermore, the temporal breadth of the dataset facilitates studies on the evolution of judicial reasoning, the influence of precedent, and shifts in legal interpretation across centuries.

Beyond argument mining, LAMUS also supports tasks such as case summarization, legal question answering, and predictive modeling of judicial decisions. Its availability as an open, structured dataset encourages reproducibility and benchmarking in computational legal studies, fostering further advancements in the intersection of law and artificial intelligence.

\begin{figure}[ht]
    \centering
    \includegraphics[width=\columnwidth]{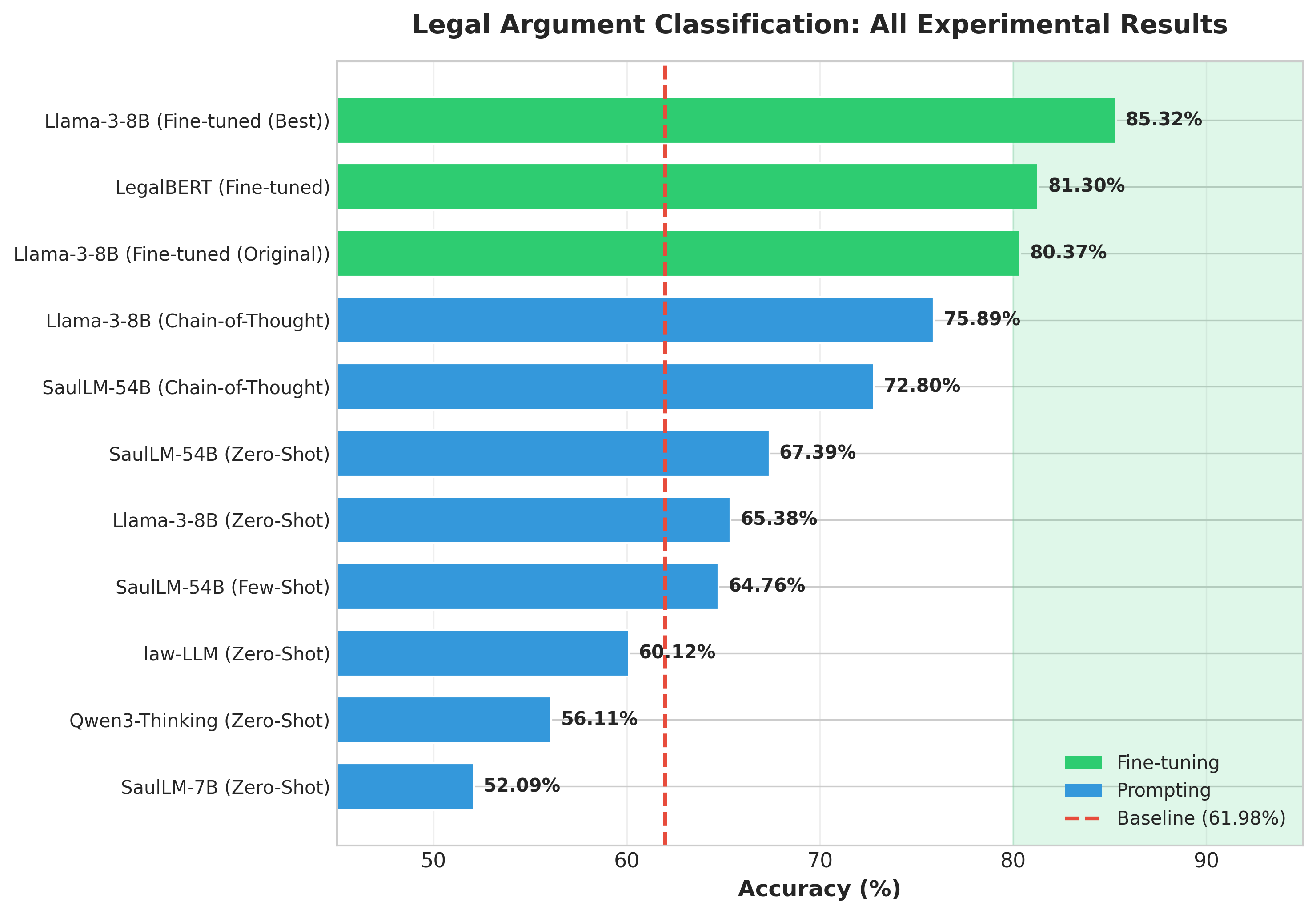}
    \caption{Overall comparison of experimental results across all models and settings, including baseline performance, accuracy percentages, and indicators for fine-tuned configurations.}
\label{fig:overall_results}
\end{figure}

\section{Conclusions}

This study evaluated the effectiveness of large language models (LLMs) for sentence-level legal argument classification, systematically exploring prompting strategies, fine-tuning, and model scales. Our experiments demonstrate that model capacity and prompting methodology critically influence classification performance, offering practical guidance for deploying LLMs in legal NLP tasks.

Based on these findings, we provide several practical recommendations for both research and deployment. For production use, fine-tuned LLaMA-3-8B is the recommended model, achieving the highest accuracy of 85.32\%. In scenarios where fine-tuning resources are limited, Chain-of-Thought prompting with LLaMA-3-8B offers a viable alternative, reaching 75.89\% accuracy. Importantly, Few-Shot prompting should be avoided for legal sentence classification tasks, as it consistently reduces model performance. These recommendations ensure that users can achieve robust results while optimizing computational resources and model selection strategies.

Figure~\ref{fig:overall_results} summarizes performance across all experimental settings, comparing zero-shot, few-shot, and fine-tuned configurations against a common baseline. The visualization highlights substantial variance in accuracy across models and prompting strategies, with fine-tuned models consistently outperforming prompt-based approaches. In contrast, few-shot prompting often fails to exceed the baseline and, in several cases, results in degraded performance relative to zero-shot evaluation. This consolidated comparison reinforces the central finding of this work: model adaptation through fine-tuning is more effective than prompt engineering alone for legal sentence classification, particularly under domain shift.

\subsection{Key Achievements}

The primary achievement of this work is the successful development of a model capable of exceeding our target accuracy for sentence-level legal argument classification.

\begin{itemize}
    \item Developed a fine-tuned LLaMA-3-8B model achieving \textbf{85.32\%} accuracy, surpassing the target of 80--85\% and improving 23.34\% over the majority-class baseline.
    \item Constructed the LAMUS corpus containing \textbf{2,900,083} sentence-level annotations from U.S. Supreme Court decisions spanning 1921--2025.
    \item Conducted 57 systematic experiments (21 prompting, 36 ablation studies) to understand performance drivers.
    \item Discovered that few-shot prompting with fixed examples \textbf{degrades} performance as example count increases (67.23\% zero-shot to 53.94\% with 100 examples), a critical negative result for legal NLP practitioners.
    \item Validated annotation quality through human verification achieving Cohen's Kappa $\kappa = 0.85$ (almost perfect agreement) with 89.2\% human-model agreement.
    \item Confirmed method stability through 10 independent runs (mean 74.71\% $\pm$ 0.56\%, $p < 0.001$).
    \item Identified Chain-of-Thought prompting as beneficial for sufficiently capable models ($\geq$8B parameters), achieving 75.89\% accuracy.
    \item Demonstrated that fine-tuning improves performance by +9.43\% over the best prompting method (85.32\% vs 75.89\% CoT).
    \item Highlighted learning rate as a sensitive hyperparameter, with variations causing up to $\pm$10\% change in accuracy.
\end{itemize}

\subsection{Future Work}

In future work, we plan to explore several directions to further extend the capabilities and applications of our framework. One promising direction is to conduct stability testing to evaluate how consistently the model predicts labels across repeated runs, providing additional insight into prediction robustness and reliability. Further experimentation with training strategies, such as curriculum learning or iterative refinement using high-quality labeled subsets, may also improve classification performance and model generalization.

Another important direction is expanding the annotation process by involving additional legal experts. Increasing expert participation can improve label consistency, reduce annotation noise, and produce higher-quality datasets that better reflect nuanced legal reasoning.

Beyond improving classification performance, future work will also investigate practical applications of legal argument classification. For example, structured argument labels could support automated legal argument generation systems by helping models identify and organize claims, evidence, and reasoning patterns in judicial decisions. Argument classification may also enable more advanced legal discourse analysis, allowing researchers to study patterns of legal reasoning across cases, courts, or jurisdictions. Additionally, these structured representations could assist legal practitioners by improving case summarization, precedent retrieval, and decision-support tools that highlight key argumentative components in legal texts. Exploring these downstream applications will help demonstrate the broader utility of legal argument classification in real-world legal AI systems.

\section*{Declaration of Generative AI and AI-assisted technologies in the writing process}
During the preparation of this work, the authors used ChatGPT in order to improve the language and grammar of the manuscript. After using this tool, the authors reviewed and edited the content as needed. The authors take full responsibility for the content of the published article.


\bibliographystyle{ACM-Reference-Format}
\bibliography{references.bib}

@article{chen2022construction,
  title={Construction and evaluation of a high-quality corpus for legal intelligence using semiautomated approaches},
  author={Chen, Haihua and Pieptea, Lavinia F and Ding, Junhua},
  journal={IEEE Transactions on Reliability},
  volume={71},
  number={2},
  pages={657--673},
  year={2022},
  publisher={IEEE}
}

@inproceedings{poudyal2020echr,
  title={ECHR: Legal corpus for argument mining},
  author={Poudyal, Prakash and {\v{S}}avelka, Jarom{\'\i}r and Ieven, Aagje and Moens, Marie Francine and Goncalves, Teresa and Quaresma, Paulo},
  booktitle={Proceedings of the 7th Workshop on Argument Mining},
  pages={67--75},
  year={2020}
}

@article{mochales2011argumentation,
  title={Argumentation mining},
  author={Mochales, Raquel and Moens, Marie-Francine},
  journal={Artificial intelligence and law},
  volume={19},
  number={1},
  pages={1--22},
  year={2011},
  publisher={Springer}
}

@article{zhang2022enhancing,
  title={Enhancing legal argument mining with domain pre-training and neural networks},
  author={Zhang, Gechuan and Nulty, Paul and Lillis, David},
  journal={Journal of Data Mining \& Digital Humanities},
  year={2022},
  publisher={Episciences. org}
}

@article{xiao2018cail2018,
  title={Cail2018: A large-scale legal dataset for judgment prediction},
  author={Xiao, Chaojun and Zhong, Haoxi and Guo, Zhipeng and Tu, Cunchao and Liu, Zhiyuan and Sun, Maosong and Feng, Yansong and Han, Xianpei and Hu, Zhen and Wang, Heng and others},
  journal={arXiv preprint arXiv:1807.02478},
  year={2018}
}

@inproceedings{chalkidis-etal-2020-legal,
    title = "{LEGAL}-{BERT}: The Muppets straight out of Law School",
    author = "Chalkidis, Ilias  and
      Fergadiotis, Manos  and
      Malakasiotis, Prodromos  and
      Aletras, Nikolaos  and
      Androutsopoulos, Ion",
    editor = "Cohn, Trevor  and
      He, Yulan  and
      Liu, Yang",
    booktitle = "Findings of the Association for Computational Linguistics: EMNLP 2020",
    month = nov,
    year = "2020",
    address = "Online",
    publisher = "Association for Computational Linguistics",
    url = "https://aclanthology.org/2020.findings-emnlp.261/",
    doi = "10.18653/v1/2020.findings-emnlp.261",
    pages = "2898--2904"
}

@article{ji2023survey,
  title={Survey of hallucination in natural language generation},
  author={Ji, Ziwei and Lee, Nayeon and Frieske, Rita and Yu, Tiezheng and Su, Dan and Xu, Yan and Ishii, Etsuko and Bang, Ye Jin and Madotto, Andrea and Fung, Pascale},
  journal={ACM computing surveys},
  volume={55},
  number={12},
  pages={1--38},
  year={2023},
  publisher={ACM New York, NY}
}

@inproceedings{cho2025modeling,
  title={Modeling Motivated Reasoning in Law: Evaluating Strategic Role Conditioning in LLM Summarization},
  author={Cho, Eunjung and Hoyle, Alexander Miserlis and Hermstr{\"u}wer, Yoan},
  booktitle={Proceedings of the Natural Legal Language Processing Workshop 2025},
  pages={68--112},
  year={2025}
}

@inproceedings{el2024factuality,
  title={The factuality of large language models in the legal domain},
  author={El Hamdani, Rajaa and Bonald, Thomas and Malliaros, Fragkiskos D and Holzenberger, Nils and Suchanek, Fabian},
  booktitle={Proceedings of the 33rd ACM International Conference on Information and Knowledge Management},
  pages={3741--3746},
  year={2024}
}

@misc{llama3,
  author       = {{Meta AI}},
  title        = {The Llama 3 Model Family},
  year         = {2024},
  howpublished = {\url{https://ai.meta.com/llama/}},
}

@article{colombo2024saullm,
  title={Saullm-54b \& saullm-141b: Scaling up domain adaptation for the legal domain},
  author={Colombo, Pierre and Pires, Telmo and Boudiaf, Malik and Melo, Rui and Culver, Dominic and Malaboeuf, Etienne and Hautreux, Gabriel and Charpentier, Johanne and Desa, Michael},
  journal={Advances in Neural Information Processing Systems},
  volume={37},
  pages={129672--129695},
  year={2024}
}

@misc{qwen3,
  author       = {{Alibaba Cloud}},
  title        = {Qwen3 Technical Report},
  year         = {2025},
  howpublished = {\url{https://qwenlm.github.io/}},
  note         = {Reasoning-oriented large language models}
}

@misc{gemini,
  author       = {{Google DeepMind}},
  title        = {Gemini Model Family},
  year         = {2025},
  howpublished = {\url{https://ai.google.dev/gemini-api}},
  note         = {Including Gemini 2.5 Flash}
}

@article{ouyang2022training,
  title={Training language models to follow instructions with human feedback},
  author={Ouyang, Long and Wu, Jeffrey and Jiang, Xu and Almeida, Diogo and Wainwright, Carroll and Mishkin, Pamela and Zhang, Chong and Agarwal, Sandhini and Slama, Katarina and Ray, Alex and others},
  journal={Advances in neural information processing systems},
  volume={35},
  pages={27730--27744},
  year={2022}
}

@article{radford2019language,
  title={Language models are unsupervised multitask learners},
  author={Radford, Alec and Wu, Jeffrey and Child, Rewon and Luan, David and Amodei, Dario and Sutskever, Ilya and others},
  journal={OpenAI blog},
  volume={1},
  number={8},
  pages={9},
  year={2019}
}

@article{hendrycks2020measuring,
  title={Measuring massive multitask language understanding},
  author={Hendrycks, Dan and Burns, Collin and Basart, Steven and Zou, Andy and Mazeika, Mantas and Song, Dawn and Steinhardt, Jacob},
  journal={arXiv preprint arXiv:2009.03300},
  year={2020}
}

@article{min2022rethinking,
  title={Rethinking the role of demonstrations: What makes in-context learning work?},
  author={Min, Sewon and Lyu, Xinxi and Holtzman, Ari and Artetxe, Mikel and Lewis, Mike and Hajishirzi, Hannaneh and Zettlemoyer, Luke},
  journal={arXiv preprint arXiv:2202.12837},
  year={2022}
}

@article{liu2023pre,
  title={Pre-train, prompt, and predict: A systematic survey of prompting methods in natural language processing},
  author={Liu, Pengfei and Yuan, Weizhe and Fu, Jinlan and Jiang, Zhengbao and Hayashi, Hiroaki and Neubig, Graham},
  journal={ACM computing surveys},
  volume={55},
  number={9},
  pages={1--35},
  year={2023},
  publisher={ACM New York, NY}
}

@article{wei2022chain,
  title={Chain-of-thought prompting elicits reasoning in large language models},
  author={Wei, Jason and Wang, Xuezhi and Schuurmans, Dale and Bosma, Maarten and Xia, Fei and Chi, Ed and Le, Quoc V and Zhou, Denny and others},
  journal={Advances in neural information processing systems},
  volume={35},
  pages={24824--24837},
  year={2022}
}

@article{kojima2022large,
  title={Large language models are zero-shot reasoners},
  author={Kojima, Takeshi and Gu, Shixiang Shane and Reid, Machel and Matsuo, Yutaka and Iwasawa, Yusuke},
  journal={Advances in neural information processing systems},
  volume={35},
  pages={22199--22213},
  year={2022}
}

@inproceedings{zhou2023least,
  title={Least-to-Most Prompting Enables Complex Reasoning in Large Language Models},
  author={Zhou, Dan and Tang, Zhijing and Sun, Huan and Chen, Weizhe and others},
  booktitle={ICLR},
  year={2023}
}

@inproceedings{bender2021dangers,
  title={On the dangers of stochastic parrots: Can language models be too big?},
  author={Bender, Emily M and Gebru, Timnit and McMillan-Major, Angelina and Shmitchell, Shmargaret},
  booktitle={Proceedings of the 2021 ACM conference on fairness, accountability, and transparency},
  pages={610--623},
  year={2021}
}

@article{chen2021data,
  title={Data evaluation and enhancement for quality improvement of machine learning},
  author={Chen, Haihua and Chen, Jiangping and Ding, Junhua},
  journal={IEEE Transactions on Reliability},
  volume={70},
  number={2},
  pages={831--847},
  year={2021},
  publisher={IEEE}
}

@article{zhang2025thinking,
  title={Thinking Longer, Not Always Smarter: Evaluating LLM Capabilities in Hierarchical Legal Reasoning},
  author={Zhang, Li and Grabmair, Matthias and Gray, Morgan and Ashley, Kevin},
  journal={arXiv preprint arXiv:2510.08710},
  year={2025}
}

@article{guha2023legalbench,
  title={Legalbench: A collaboratively built benchmark for measuring legal reasoning in large language models},
  author={Guha, Neel and Nyarko, Julian and Ho, Daniel and R{\'e}, Christopher and Chilton, Adam and Chohlas-Wood, Alex and Peters, Austin and Waldon, Brandon and Rockmore, Daniel and Zambrano, Diego and others},
  journal={Advances in neural information processing systems},
  volume={36},
  pages={44123--44279},
  year={2023}
}

@article{zha2025data,
  title={Data-centric artificial intelligence: A survey},
  author={Zha, Daochen and Bhat, Zaid Pervaiz and Lai, Kwei-Herng and Yang, Fan and Jiang, Zhimeng and Zhong, Shaochen and Hu, Xia},
  journal={ACM Computing Surveys},
  volume={57},
  number={5},
  pages={1--42},
  year={2025},
  publisher={ACM New York, NY}
}

@article{gilardi2023chatgpt,
  title={ChatGPT outperforms crowd workers for text-annotation tasks},
  author={Gilardi, Fabrizio and Alizadeh, Meysam and Kubli, Ma{\"e}l},
  journal={Proceedings of the National Academy of Sciences},
  volume={120},
  number={30},
  pages={e2305016120},
  year={2023},
  publisher={National Academy of Sciences}
}

@article{guo2025specialized,
  title={Specialized or general AI? a comparative evaluation of LLMs’ performance in legal tasks},
  author={Guo, Xue and Huang, Yuting and Wei, Bin and Kuang, Kun and Wu, Yiquan and Gan, Leilei and Huang, Xianshan and Dong, Xianglin},
  journal={Artificial Intelligence and Law},
  pages={1--37},
  year={2025},
  publisher={Springer}
}

@article{li2024llms,
  title={Llms-as-judges: a comprehensive survey on llm-based evaluation methods},
  author={Li, Haitao and Dong, Qian and Chen, Junjie and Su, Huixue and Zhou, Yujia and Ai, Qingyao and Ye, Ziyi and Liu, Yiqun},
  journal={arXiv preprint arXiv:2412.05579},
  year={2024}
}

@article{shao2025large,
  title={When large language models meet law: Dual-lens taxonomy, technical advances, and ethical governance},
  author={Shao, Peizhang and Xu, Linrui and Wang, Jinxi and Zhou, Wei and Wu, Xingyu},
  journal={arXiv preprint arXiv:2507.07748},
  year={2025}
}

@article{chi2026legalai,
  title={LegalAi research in LLM Era: data, modeling and evaluation},
  author={Chi, Xiao and Wang, Wei and Zhang, Ziyao and Li, Ang and Huang, Yuting and Wu, Yiquan and Kuang, Kun and Sun, Changlong and Liu, Xiaozhong and Wu, Fei and others},
  journal={Artificial Intelligence Review},
  year={2026},
  publisher={Springer}
}

@article{hu2026evaluation,
  title={Evaluation of Large Language Models in Legal Applications: Challenges, Methods, and Future Directions},
  author={Hu, Yiran and Liu, Huanghai and Wang, Chong and Li, Kunran and Wu, Tien-Hsuan and Li, Haitao and Xu, Xinran and Huo, Siqing and Su, Weihang and Zheng, Ning and others},
  journal={arXiv preprint arXiv:2601.15267},
  year={2026}
}

@article{landis1977measurement,
  title={The measurement of observer agreement for categorical data},
  author={Landis, J Richard and Koch, Gary G},
  journal={biometrics},
  pages={159--174},
  year={1977},
  publisher={JSTOR}
}

@incollection{bommarito2021lexnlp,
  title={LexNLP: Natural language processing and information extraction for legal and regulatory texts},
  author={Bommarito II, Michael J and Katz, Daniel Martin and Detterman, Eric M},
  booktitle={Research handbook on big data law},
  pages={216--227},
  year={2021},
  publisher={Edward Elgar Publishing}
}

@inproceedings{palau2009argumentation,
  title={Argumentation mining: the detection, classification and structure of arguments in text},
  author={Palau, Raquel Mochales and Moens, Marie-Francine},
  booktitle={Proceedings of the 12th international conference on artificial intelligence and law},
  pages={98--107},
  year={2009}
}

@article{ariai2025natural,
  title={Natural language processing for the legal domain: A survey of tasks, datasets, models, and challenges},
  author={Ariai, Farid and Mackenzie, Joel and Demartini, Gianluca},
  journal={ACM Computing Surveys},
  volume={58},
  number={6},
  pages={1--37},
  year={2025},
  publisher={ACM New York, NY}
}

@article{liga2023fine,
  title={Fine-tuning GPT-3 for legal rule classification},
  author={Liga, Davide and Robaldo, Livio},
  journal={Computer Law \& Security Review},
  volume={51},
  pages={105864},
  year={2023},
  publisher={Elsevier}
}

@inproceedings{li2025legalagentbench,
  title={Legalagentbench: Evaluating llm agents in legal domain},
  author={Li, Haitao and Chen, Junjie and Yang, Jingli and Ai, Qingyao and Jia, Wei and Liu, Youfeng and Lin, Kai and Wu, Yueyue and Yuan, Guozhi and Hu, Yiran and others},
  booktitle={Proceedings of the 63rd Annual Meeting of the Association for Computational Linguistics (Volume 1: Long Papers)},
  pages={2322--2344},
  year={2025}
}

@article{siino2025exploring,
  title={Exploring llms applications in law: A literature review on current legal nlp approaches},
  author={Siino, Marco and Falco, Mariana and Croce, Daniele and Rosso, Paolo},
  journal={IEEE Access},
  volume={13},
  pages={18253--18276},
  year={2025},
  publisher={IEEE}
}

@inproceedings{enguehard2025lemaj,
  title={LeMAJ (Legal LLM-as-a-Judge): Bridging Legal Reasoning and LLM Evaluation},
  author={Enguehard, Joseph and Van Ermengem, Morgane and Atkinson, Kate and Cha, Sujeong and Chowdhury, Arijit Ghosh and Ramaswamy, Prashanth Kallur and Roghair, Jeremy and Marlowe, Hannah R and Negreanu, Carina Suzana and Boxall, Kitty and others},
  booktitle={Proceedings of the Natural Legal Language Processing Workshop 2025},
  pages={318--337},
  year={2025}
}

@inproceedings{shu2024lawllm,
  title={Lawllm: Law large language model for the us legal system},
  author={Shu, Dong and Zhao, Haoran and Liu, Xukun and Demeter, David and Du, Mengnan and Zhang, Yongfeng},
  booktitle={Proceedings of the 33rd ACM International Conference on information and knowledge management},
  pages={4882--4889},
  year={2024}
}

@article{kant2025towards,
  title={Towards robust legal reasoning: Harnessing logical llms in law},
  author={Kant, Manuj and Nabi, Sareh and Kant, Manav and Scharrer, Roland and Ma, Megan and Nabi, Marzieh},
  journal={arXiv preprint arXiv:2502.17638},
  year={2025}
}

\appendix
\section{Appendix}

This appendix provides comprehensive supplementary results from our experimental evaluation. We present the complete ablation grid covering all 36 hyperparameter configurations, stability test results demonstrating method reproducibility, a complete ranking of all evaluated models, extended few-shot analysis with additional metrics, per-class performance comparisons, domain shift quantification, and reference scales for interpreting agreement statistics. These supplementary materials enable full reproducibility of our findings and provide additional insights beyond the main text.


\subsection{Complete Ablation Grid}

To systematically identify optimal hyperparameters for fine-tuning LLaMA-3-8B on legal argument classification, we conducted a comprehensive grid search across 36 configurations. Table~\ref{table:ablation_full} presents accuracy results for all combinations of four learning rates (1e-5, 5e-5, 1e-4, 2e-4), three LoRA ranks (8, 16, 32), and three training durations (1, 3, 5 epochs).

\begin{table*}[h]
\caption{Complete ablation grid results for fine-tuned LLaMA-3-8B. All values are test accuracy (\%). Best result: 85.32\% (LR=1e-4, Rank=8, Epochs=5).}
\begin{center}
\begin{tabular}{c|ccc|ccc|ccc}
\hline
 & \multicolumn{3}{c|}{\textbf{Epochs = 1}} & \multicolumn{3}{c|}{\textbf{Epochs = 3}} & \multicolumn{3}{c}{\textbf{Epochs = 5}} \\
\textbf{LR} & R=8 & R=16 & R=32 & R=8 & R=16 & R=32 & R=8 & R=16 & R=32 \\
\hline
1e-5 & 55.95 & 66.62 & 69.24 & 70.02 & 75.12 & 79.60 & 76.82 & 79.75 & 81.76 \\
5e-5 & 74.96 & 78.83 & 81.30 & 82.23 & 83.15 & 83.93 & 84.23 & 84.70 & 85.01 \\
1e-4 & 79.44 & 80.68 & 81.76 & 83.77 & 84.54 & 84.23 & \textbf{85.32} & \textbf{85.32} & 85.01 \\
2e-4 & 81.30 & 82.69 & 83.62 & 84.70 & 85.16 & 84.39 & 84.85 & 83.93 & 83.31 \\
\hline
\end{tabular}
\end{center}
\label{table:ablation_full}
\end{table*}

Several patterns emerge from the ablation results. First, learning rate exhibits the strongest influence on performance: at 1 epoch, accuracy ranges from 55.95\% (LR=1e-5, R=8) to 83.62\% (LR=2e-4, R=32), a spread of nearly 28 percentage points. Second, the optimal learning rate shifts with training duration---higher learning rates (2e-4) perform best for shorter training (1--3 epochs), while moderate rates (1e-4) excel with longer training (5 epochs), likely because aggressive learning rates cause overfitting when combined with extended training. Third, LoRA rank has minimal impact once learning rate and epochs are optimized, with accuracy varying by less than 1\% across ranks at optimal settings. This suggests that rank 8 is sufficient for this classification task, reducing computational overhead without sacrificing accuracy. The best configuration (LR=1e-4, Rank=8, Epochs=5) achieves 85.32\% accuracy, matching the result obtained with Rank=16 at the same settings.


\subsection{Stability Test Results}

To assess the reproducibility of our prompting-based results, we conducted stability testing by running Chain-of-Thought prompting with LLaMA-3-8B across 10 independent trials using different random seeds. Table~\ref{table:stability} reports the accuracy achieved in each run.

\begin{table}[hbt!]
\caption{Stability test results for Chain-of-Thought prompting with LLaMA-3-8B across 10 independent runs. Mean accuracy: 74.71\%, standard deviation: 0.56\%.}
\begin{center}
\begin{tabular}{ccc}
\hline
\textbf{Run} & \textbf{Seed} & \textbf{Accuracy (\%)} \\
\hline
1 & 142 & 75.12 \\
2 & 242 & 75.43 \\
3 & 342 & 74.50 \\
4 & 442 & 74.34 \\
5 & 542 & 74.96 \\
6 & 642 & 74.34 \\
7 & 742 & 74.65 \\
8 & 842 & 75.58 \\
9 & 942 & 73.72 \\
10 & 1042 & 74.50 \\
\hline
\textbf{Mean} & -- & \textbf{74.71} \\
\textbf{Std Dev} & -- & \textbf{0.56} \\
\hline
\end{tabular}
\end{center}
\label{table:stability}
\end{table}

The results demonstrate high reproducibility, with a standard deviation of only 0.56\% across runs. Individual accuracies ranged from 73.72\% to 75.58\%, representing a spread of less than 2 percentage points. A one-sample t-test confirmed that the mean accuracy (74.71\%) significantly exceeds the zero-shot baseline (67.23\%) with $p < 0.001$, validating that the performance gains from Chain-of-Thought prompting are statistically robust rather than artifacts of random variation. This stability supports the reliability of our comparative findings and suggests that practitioners can expect consistent performance when deploying CoT prompting for legal argument classification.


\subsection{Complete Model Comparison}

Table~\ref{table:all_results} provides a comprehensive ranking of all model-method combinations evaluated in this study, ordered by accuracy. This complete view contextualizes the relative performance of different approaches and highlights the magnitude of improvement achieved through fine-tuning.

\begin{table}[h]
\caption{Complete results across all models and methods, ranked by accuracy. Baseline (majority class): 61.98\%. The $\Delta$ column shows improvement over baseline.}
\begin{center}
\small
\begin{tabular}{llcc}
\hline
\textbf{Model} & \textbf{Method} & \textbf{Acc.} & \textbf{$\Delta$} \\
\hline
LLaMA-3-8B & Fine-tuned (Best) & \textbf{85.32\%} & +23.34 \\
LLaMA-3-8B & Fine-tuned (2e-4) & 85.16\% & +23.18 \\
LegalBERT & Fine-tuned & 81.30\% & +19.32 \\
LLaMA-3-8B & Fine-tuned (Orig) & 80.37\% & +18.39 \\
LLaMA-3-8B & Chain-of-Thought & 75.89\% & +13.91 \\
SaulLM-54B & Chain-of-Thought & 72.80\% & +10.82 \\
SaulLM-54B & Zero-Shot & 67.39\% & +5.41 \\
LLaMA-3-8B & Zero-Shot & 65.38\% & +3.40 \\
SaulLM-54B & Few-Shot & 64.76\% & +2.78 \\
law-LLM & Zero-Shot & 60.12\% & -1.86 \\
Qwen3-Thinking & Zero-Shot & 56.11\% & -5.87 \\
Qwen3-Thinking & CoT & 54.10\% & -7.88 \\
SaulLM-7B & Zero-Shot & 52.09\% & -9.89 \\
Qwen3-Thinking & Few-Shot & 49.30\% & -12.68 \\
LLaMA-3-8B & Few-Shot & 45.75\% & -16.23 \\
SaulLM-7B & CoT & 38.02\% & -23.96 \\
law-LLM & Few-Shot & 31.68\% & -30.30 \\
law-LLM & CoT & 28.75\% & -33.23 \\
SaulLM-7B & Few-Shot & 21.64\% & -40.34 \\
Gemini-2.5 & All & 5.41\% & -56.57 \\
\hline
\end{tabular}
\end{center}
\label{table:all_results}
\end{table}

The ranking reveals several important patterns. Fine-tuned models occupy the top four positions, with accuracies ranging from 80.37\% to 85.32\%, demonstrating the substantial advantage of task-specific adaptation. Among prompting approaches, Chain-of-Thought with LLaMA-3-8B (75.89\%) significantly outperforms all other prompted configurations. Notably, few-shot prompting consistently underperforms zero-shot baselines for most models, with LLaMA-3-8B few-shot (45.75\%) falling nearly 20 percentage points below its zero-shot result (65.38\%). Several smaller models (SaulLM-7B, law-LLM with CoT/Few-Shot) perform below the majority-class baseline, indicating that neither domain specialization nor advanced prompting compensates for insufficient model capacity. Gemini-2.5-Flash produced near-random outputs (5.41\%) due to parsing failures, highlighting the importance of output format compatibility in LLM evaluation.


\subsection{Extended Few-Shot Analysis}

To thoroughly investigate the relationship between example count and classification accuracy, we tested LLaMA-3-8B with 0 to 100 few-shot examples. Table~\ref{table:fewshot_full} presents accuracy on both the full 6-class task and a reduced 5-class variant (excluding ``Others'').

\begin{table}[h]
\caption{Extended few-shot sweep (0--100 examples) for LLaMA-3-8B. Accuracy is reported for both 6-class and 5-class (excluding ``Others'') settings. Performance degrades monotonically as example count increases.}
\begin{center}
\begin{tabular}{cccc}
\hline
\textbf{\# Ex.} & \textbf{Acc. (6-class)} & \textbf{Acc. (5-class)} & \textbf{$\Delta$ ZS} \\
\hline
0 & 67.23\% & 71.08\% & baseline \\
5 & 65.07\% & 68.79\% & -2.16\% \\
10 & 66.15\% & 69.93\% & -1.08\% \\
20 & 64.91\% & 68.63\% & -2.32\% \\
40 & 65.53\% & 69.28\% & -1.70\% \\
60 & 60.43\% & 63.89\% & -6.80\% \\
80 & 59.04\% & 62.42\% & -8.19\% \\
100 & 53.94\% & 57.03\% & -13.29\% \\
\hline
\end{tabular}
\end{center}
\label{table:fewshot_full}
\end{table}

The results reveal a counterintuitive finding: increasing the number of few-shot examples consistently degrades performance. Accuracy drops from 67.23\% with zero examples to 53.94\% with 100 examples---a decrease of over 13 percentage points. This pattern holds for both 6-class and 5-class settings. We attribute this degradation to domain mismatch: the curated few-shot examples represent generic legal scenarios, while the test data consists of jurisdiction-specific Texas criminal court language with distinct terminology and argumentative patterns. As more mismatched examples are provided, the model increasingly overfits to irrelevant patterns in the demonstrations rather than learning generalizable classification criteria. This finding has important practical implications: for domain-specific legal classification tasks, generic few-shot examples may harm rather than help performance, and zero-shot or fine-tuned approaches should be preferred.


\subsection{Per-Class Performance}

Table~\ref{table:perclass} compares F1-scores across argument categories for the three best-performing approaches: fine-tuned LLaMA-3-8B, fine-tuned LegalBERT, and Chain-of-Thought prompting.

\begin{table}[h]
\caption{Per-class F1-scores for top-performing methods. Fine-tuned models achieve more balanced performance across categories, while CoT prompting shows greater variance.}
\begin{center}
\begin{tabular}{lccc}
\hline
\textbf{Category} & \textbf{LLaMA-FT} & \textbf{LegalBERT} & \textbf{CoT} \\
\hline
Facts & 0.91 & 0.92 & 0.82 \\
Issue & 0.73 & 0.78 & 0.65 \\
Rule/Law/Holding & 0.71 & 0.46 & 0.58 \\
Analysis & 0.54 & 0.61 & 0.48 \\
Conclusion & 0.69 & 0.71 & 0.62 \\
Others & 0.53 & 0.51 & 0.41 \\
\hline
\textbf{Weighted Avg} & 0.80 & 0.81 & 0.77 \\
\textbf{Macro Avg} & 0.69 & 0.66 & 0.59 \\
\hline
\end{tabular}
\end{center}
\label{table:perclass}
\end{table}

All methods perform best on ``Facts,'' the majority class, with F1-scores exceeding 0.82. Performance drops substantially for minority classes: ``Analysis'' and ``Others'' prove most challenging, with F1-scores below 0.61 across all methods. Interestingly, fine-tuned LLaMA-3-8B achieves the highest macro-average F1 (0.69), indicating more balanced performance across classes despite LegalBERT's marginally higher weighted average (0.81 vs 0.80). LegalBERT particularly struggles with ``Rule/Law/Holding'' (F1=0.46), possibly due to confusion with ``Analysis'' sentences that also reference legal rules. Chain-of-Thought prompting shows the largest performance gap between majority and minority classes, suggesting that explicit reasoning helps most for common patterns but provides less benefit for rare argument types. These results highlight that class imbalance remains a significant challenge for legal argument classification, and future work should explore techniques such as class-weighted loss functions or data augmentation for minority categories.


\subsection{Domain Shift Analysis}

Table~\ref{table:domain_shift_appendix} quantifies the distributional differences between the Texas Criminal Cases training corpus and the SCOTUS target corpus.

\begin{table}[h]
\caption{Domain shift between Texas Criminal Cases (training) and SCOTUS (target) corpora. Positive shifts indicate categories more prevalent in SCOTUS; negative shifts indicate categories more prevalent in Texas data.}
\begin{center}
\begin{tabular}{lccc}
\hline
\textbf{Category} & \textbf{Texas (\%)} & \textbf{SCOTUS (\%)} & \textbf{Shift} \\
\hline
Facts & 61.9 & 26.3 & -35.6 \\
Issue & 4.2 & 2.1 & -2.1 \\
Rule/Law/Holding & 5.3 & 27.6 & +22.2 \\
Analysis & 14.3 & 27.6 & +13.3 \\
Conclusion & 8.9 & 4.2 & -4.7 \\
Others & 5.4 & 12.2 & +6.9 \\
\hline
\end{tabular}
\end{center}
\label{table:domain_shift_appendix}
\end{table}

The analysis reveals substantial domain shift between the two corpora. The Texas training data is heavily dominated by ``Facts'' (61.9\%), reflecting the emphasis on establishing evidentiary foundations in state criminal appeals. In contrast, SCOTUS opinions exhibit a more balanced distribution, with ``Rule/Law/Holding'' (27.6\%) and ``Analysis'' (27.6\%) together comprising over half of all sentences, consistent with the Supreme Court's focus on interpreting and applying legal doctrine rather than establishing factual records. The 35.6 percentage-point decrease in ``Facts'' and corresponding increases in rule-based and analytical content highlight fundamental differences in argumentative structure between state criminal appeals and federal constitutional review. This domain shift poses challenges for models trained on Texas data when applied to SCOTUS opinions, and explains why robust performance requires either fine-tuning on representative target-domain data or prompting strategies that generalize across distributional differences.


\subsection{Cohen's Kappa Interpretation}

Table~\ref{table:kappa_scale} provides the standard interpretation scale for Cohen's Kappa, enabling readers to contextualize our inter-annotator agreement results.

\begin{table}[h]
\caption{Cohen's Kappa interpretation scale following Landis \& Koch (1977). Our human verification achieved $\kappa = 0.85$, indicating almost perfect agreement.}
\begin{center}
\begin{tabular}{lc}
\hline
\textbf{Kappa Range} & \textbf{Interpretation} \\
\hline
0.81 -- 1.00 & Almost Perfect $\leftarrow$ \textbf{Ours: 0.85} \\
0.61 -- 0.80 & Substantial \\
0.41 -- 0.60 & Moderate \\
0.21 -- 0.40 & Fair \\
0.00 -- 0.20 & Slight \\
\hline
\end{tabular}
\end{center}
\label{table:kappa_scale}
\end{table}

Cohen's Kappa ($\kappa$) measures inter-rater agreement while correcting for agreement expected by chance, making it more informative than simple percent agreement for classification tasks with imbalanced categories. Our achieved $\kappa = 0.85$ falls within the ``Almost Perfect'' range (0.81--1.00), indicating that human annotators and the LLM-based labeling pipeline produce highly consistent annotations. This level of agreement exceeds thresholds commonly considered acceptable for corpus construction in NLP research ($\kappa \geq 0.67$) and legal annotation studies ($\kappa \geq 0.70$). The high agreement validates both the clarity of our annotation guidelines and the reliability of LLaMA-3-70B for automated sentence-level legal argument labeling, supporting the use of the LAMUS corpus for downstream research and model development.


\end{document}